\documentclass[aps,10pt,notitlepage,superscriptaddress,pra,twocolumn]{revtex4-1}

\usepackage{amsmath}
\usepackage{amsfonts}
\usepackage{amssymb}
\usepackage{graphicx}
\usepackage{sidecap}
\usepackage{xcolor}
\usepackage{braket}

\usepackage{blindtext}
\usepackage{lipsum}
\newcommand*\diff{\mathop{}\!\mathrm{d}}
\usepackage{hyperref}
\usepackage{mathtools}
\renewcommand{\vec}[1]{\boldsymbol{#1}}

\usepackage{enumitem}
\usepackage{appendix}
\usepackage{subfigure}
\usepackage{color}
\usepackage[dvipsnames]{xcolor}
\usepackage{physics}
\usepackage{hyperref}
\hypersetup{
	colorlinks=true,       
	linkcolor=VioletRed,  
    citecolor=JungleGreen,        
    filecolor=Orchid,      
    urlcolor=black           
}

 \usepackage{amsthm}

\newtheorem*{definition}{Definition}
\newtheorem*{assumption}{Assumption}

\begin{document}
\title{
Ghost mechanism: An analytical model of abrupt learning in recurrent networks
}

\author{Fatih Dinc}
\thanks{Core contributors}
\affiliation{Kavli Institute for Theoretical Physics, University of California, Santa Barbara, CA 93106, USA}
\affiliation{Geometric Intelligence Lab, University of California, Santa Barbara, CA 93106, USA}
\affiliation{CNC Program, Stanford University, Stanford, CA 94305, USA}
\affiliation{Physics of Artificial Intelligence Group, NTT Research Inc., Sunnyvale, CA 94085, USA}
\author{Ege Cirakman}
\thanks{Core contributors}
\affiliation{CNC Program, Stanford University, Stanford, CA 94305, USA}
\author{Bariscan Kurtkaya}
\thanks{Core contributors}
\affiliation{CNC Program, Stanford University, Stanford, CA 94305, USA}
\affiliation{KUIS AI, Department of Computer Engineering, Koc University, Istanbul, Turkey}
\author{Mert Yuksekgonul}
\affiliation{Computer Science, Stanford University, Stanford, CA 94305, USA}
\author{Yiqi Jiang}
\affiliation{CNC Program, Stanford University, Stanford, CA 94305, USA}
\author{Mark Schnitzer}
\thanks{Co-supervision}
\affiliation{CNC Program, Stanford University, Stanford, CA 94305, USA}
\affiliation{James H. Clark Center for Biomedical Engineering \& Sciences, Stanford University, Stanford}
\affiliation{Howard Hughes Medical Institute, Stanford University, Stanford, CA 94305, USA}
\author{Hidenori Tanaka}
\thanks{Co-supervision}
\affiliation{Physics of Artificial Intelligence Group, NTT Research Inc., Sunnyvale, CA 94085, USA}
\affiliation{CBS-NTT Program in Physics of Intelligence, Harvard University, Cambridge, MA 94305, USA}

\begin{abstract}
\textit{Abrupt learning}, long performance plateaus followed by rapid convergence, is a common phenomenon in recurrent neural networks (RNNs) trained on working memory tasks. In such cases, the networks develop transient slow regions in state space that extend the effective timescales of computation. However, the mechanisms driving sudden performance improvements and their causal role remain unclear, largely because we lack an analytical dynamical-systems framework. To address this gap, we introduce the ghost mechanism, a general process by which finite-dimensional continuous-time dynamical systems exhibit transient slowdown near the remnant of a saddle-node bifurcation. By reducing the high-dimensional dynamics near ghost points, we derive a one-dimensional canonical form that analytically captures learning as a process controlled by a single scale parameter. Using this model, we study a form of abrupt learning emerging from ghost points and identify a critical learning rate that scales as an inverse power law with the timescale of the learned computation. Beyond this rate, learning collapses through two interacting modes: (i) vanishing gradients and (ii) oscillatory gradients near minima. These features can lock the system into high-confidence but incorrect predictions when parameter updates trigger a no-learning zone, a region of parameter space where gradients vanish. We validate these predictions in low-rank RNNs, where ghost points precede abrupt transitions, and further demonstrate their generality in full-rank RNNs trained on canonical working memory tasks. Our theory offers two approaches to address these learning difficulties: increasing trainable ranks stabilizes learning trajectories, while reducing output confidence mitigates entrapment in no-learning zones. Overall, the ghost mechanism reveals how the computational demands of a task constrain the optimization landscape, demonstrating that well-known learning difficulties in RNNs partly arise from the dynamical systems they must learn to implement.
\end{abstract}

\maketitle

\section{Introduction}
A hallmark of natural and artificial neural networks is the \emph{emergence} of new capabilities, where computational and behavioral algorithms are organically learned from experience~\cite{hebb2005organization, kandel2000principles, urai2022large, wright2022deep, mcmahon2023physics, pai2023experimentally, hassabis2017neuroscience, zador2019critique}. Empirically, such an emergence often manifests itself as \emph{abrupt transitions} in learning dynamics, sudden drops in loss functions that mark the acquisition of new computational abilities~\cite{eisenmann2023bifurcations, haputhanthri2024why, delmastro2023dynamics, srivastava2022beyond, arora2023theory, yu2023skill, lubana2024percolation, okawa2023compositional, wei2022emergent, hoffmann2022empirical, power2022grokking}.  Such abrupt transitions are frequently accompanied by instabilities, including exploding or vanishing gradients and sharp shifts in network behavior \cite{pascanu2013difficulty, doya1992bifurcations, goodfellow2014qualitatively}.  While numerous studies have documented slow learning followed by sudden performance jumps \cite{reddy2023mechanistic,olsson2022context}, the underlying mechanisms driving these emergent transitions, even within specific domains of networks or classes of tasks, remain poorly understood. In this work, we perform one such mechanistic study for recurrent neural networks (RNNs) trained to perform working memory tasks.

One key challenge in studying the causal mechanisms subserving abrupt learning is that most approaches remain tied to numerical analyses \cite{sussillo2009generating,haputhanthri2024why}, particular network architectures \cite{eisenmann2023bifurcations,haputhanthri2024why,hess2023generalized}, or simplified linear models \cite{lewkowycz2020large,bahri2020statistical}, making it difficult to identify general principles. Yet, at least intuitively, one instance of universality is expected if one focuses on temporal problems solved by dynamical systems: these systems must somehow reorganize their internal dynamics to suddenly perform new computations. This suggests that rather than cataloging emergent behaviors across different systems, we might instead search for universal mechanisms, \textit{i.e.}, motifs in solutions that different architectures tend to converge to. 

To understand what these universal mechanisms might be, it is instructive to examine how different research communities have approached abrupt learning. In modern deep networks, abrupt drops in loss with step-like gains in capability are often described as emergent phenomena, especially at large model, data, and compute scales~\cite{wei2022emergent, power2022grokking, lubana2024percolation, okawa2023compositional}. While these studies provide extensive documentation of when and where such transitions occur, most accounts remain tied to specific architectural assumptions \cite{wei2022emergent} and rely on empirical evidence without identifying the underlying dynamical motifs \cite{kangaslahti2024loss}. This empirical focus, though valuable for cataloging the phenomenon, does not reveal whether different systems achieve abrupt learning through shared mechanisms.

By contrast, dynamical systems theory offers tools to identify such universal mechanisms, though traditionally studied mostly in the context of few-dimensional dynamical systems \cite{strogatz2018nonlinear}. Then, sharp performance shifts do not need to depend on the particular architecture being trained, rather on the general properties of (low-dimensional) dynamical systems approximated by these architectures \cite{schafer2006recurrent}, whereas properties of the learned dynamical systems can be analyzed using established mathematical frameworks~\cite{carr2012applications}. For instance, wisdom from classical bifurcation theory \footnote{Bifurcations are formally defined as shifts in system behavior, which alter the number or stability of fixed points, periodic orbits, or other invariant sets~\cite{strogatz2018nonlinear, schmidt2021identifying, HASCHKE200525, rehmer2022, avrutin2012occurrence, ganguli2005dangerous, hess2023generalized, Monfared2020}} would suggest that abrupt learning can occur when training drives the system across qualitative thresholds in its flow \cite{eisenmann2023bifurcations}, suggesting that bifurcations might be one such universal motif underlying sudden behavioral changes ~\cite{strogatz2018nonlinear,eisenmann2023bifurcations}. This mathematical perspective shifts focus from architecture-specific details to dynamical structures that are learned during training. In fact, these agnostic approaches have proven particularly valuable in computational neuroscience, where analyses of RNNs have identified bifurcations as key drivers of both learning instabilities \cite{pascanu2013difficulty,doya1992bifurcations} and the emergence of new computational capabilities \cite{eisenmann2023bifurcations,delmastro2023dynamics}. However, since both the parameters and the state variables are high-dimensional in the trained architectures, the precise causal mechanisms underlying such phenomena often remain poorly understood.

In this work, we identify one such universal mechanism underlying abrupt learning in dynamical system models trained to perform working memory tasks: the formation of \textit{ghost points}, \textit{i.e.}, transient bottlenecks in the dynamics of state variables that emerge near the remnants of saddle-node bifurcations \cite{strogatz2018nonlinear}. Our analysis reveals how ghosts emerge through learning, why they cause training instabilities, and how these instabilities can be mitigated. Beyond providing one analytically tractable model of abrupt learning, this work discusses testable predictions for biological neural networks, practical strategies for training recurrent networks effectively, and potential insights for learning in deep neural networks.

We develop this analysis systematically as follows. In Section~\ref{sec:theory}, we introduce an analytically tractable one-dimensional canonical model of the ghost mechanism, trained on a delayed-activation task. This model captures a form of abrupt learning as a transient slowdown in the dynamics of trained systems. Next, in section \ref{sec:lowrank}, we show that the ghost mechanism reproduces the behavior of rank-one and rank-two RNNs trained on working memory tasks, including their training instabilities. Section~\ref{sec:results} extends these insights to full-rank RNNs, where we identify two strategies to stabilize learning: (i) reducing output confidence and (ii) increasing the number of trainable ranks. {In our Discussion (section~\ref{sec:discussion}),} we situate our results within the {broader literature on machine learning and computational neuroscience}, explain how the uncovered failure modes (no-learning zones) differ from vanishing gradient problems in feedforward or recurrent neural networks that arise from saturated neuronal activation functions \cite{pascanu2013difficulty,bengio1994learning,bahri2020statistical}, and discuss their implications for auto-regressive deep learning architectures that must learn long-term dependencies without explicit mechanisms such as attention \cite{vaswani2017attention}. Finally, we conclude that addressing these constraints on training dynamical systems may shed light on learning in biological neural networks.

\section{Ghost mechanism} \label{sec:theory}
In this section we introduce the ghost mechanism and motivate its broader relevance for dynamical systems. To start with, we derive a mathematical reduction of ghosts from high-dimensional systems to a one-dimensional system that will serve as our canonical model. Using this model, we conduct analytical derivations of the learning dynamics and extract insights regarding the abrupt learning of ghost points. 

\textbf{A note on notation:} Throughout this work, we refer to scalars with lowercase letters such as $a \in \mathbb R$. We use lowercase letters with bold fonts to denote the vectors such as $\vec a \in \mathbb R^N$, a vector of $N$ element. Then, $a_i$ refers to the $i$th element of the vector. Uppercase letters with bold fonts are reserved for matrices such as $\vec A \in \mathbb R^{N \times N}$, a matrix of size $N$ by $N$. Then, $A_{ij}$ refers to the $(i,j)$th element of the matrix.

\subsection{Ghosts as temporal bottlenecks in high-dimensional dynamics}
To introduce and motivate the generality of the ghost mechanism, we first define a general finite-dimensional continuous-time dynamical system
\begin{equation} 
    \dot {\vec x}(t) = \vec f(\vec x(t)), \quad \vec x \in \mathbb{R}^N,
\end{equation}
where we refer to $\vec f:\mathbb R^N \to \mathbb R^N$ as the vector field and $\vec x$ as the state variables. We consider the behavior of this system in the neighborhood of local minima of the (kinetic) energy function,
\begin{equation}\label{eq:energy_func}
    s(\vec x) = \tfrac{1}{2}\|\vec f(\vec x)\|_2^2.
\end{equation}
We refer to a state variable configuration, in which this function attains a local minimum, as a local energy (and speed) minimum $\vec x^*$. We use a colloquial umbrella term \textit{dynamical bottleneck} to refer to a compact region around such minima. One form of local minima is called a \textit{fixed point}, for which $\vec f(\vec x^*)= \vec 0$. Not all local minima of energy function have this property, and the main goal of this subsection is to precisely define these different types of local minima, \textit{i.e.}, slow points and ghost points, for general high-dimensional dynamical systems \footnote{As a side note for precision, for our derivations in this section, we assume that $F(x(t))$ is a sufficiently smooth vector field that can be Taylor expanded to at least second order, with a non-vanishing quadratic coefficient along the slow mode direction. The exact assumption will be clear at the end of the calculation.}. In the next subsection, using these definitions, we reduce the high-dimensional dynamics to a one-dimensional variable that captures the effective behavior of the system in the neighborhood of a ghost point $\vec x^*$. Such reductions are classical in dynamical systems theory, where slow channels in high dimensions are often captured by one-dimensional normal forms (\textit{e.g.}, center manifold theory and saddle-node bifurcations~\cite{carr2012applications,kuznetsov1998elements,strogatz2018nonlinear}). This reduction will eventually allow an analytical study of learning dynamics as ghosts are formed in dynamical systems.

\begin{definition}[Slow point]
We define a slow point at a location $\vec x^*$ as one that satisfies the following conditions:
\begin{equation}
    \vec \nabla s(\vec x^*) = \vec 0, \qquad \vec \nabla^2 s(\vec x^*) \succ \vec 0, \qquad s(\vec x^*) \neq 0.
\end{equation}
Here, we define $\vec \nabla s(\vec x^*) \in \mathbb R^N$ as the gradient with respect to $x$ and $\vec \nabla^2s(\vec x^*) \in \mathbb R^{N\times N}$ as the Hessian matrix. The first condition ensures that the energy function has an extremum at $\vec x^*$, the second makes it a local minimum \footnote{To accommodate slow point manifolds, and not just isolated points, one could rework this definition. For our purposes, the distinction is not relevant for the canonical model}, and the third guarantees that the flow does not vanish at $\vec x^*$, \textit{i.e.}, creating a distinction between fixed and slow points.
\end{definition}
Following this definition, differentiating $s(\vec x)$ gives
\begin{equation}
    \vec \nabla s(\vec x) = \vec J(\vec x)^T \vec f(\vec x),
\end{equation}
where $J_{ij}=\frac{\partial f_i}{\partial x_j}$ is the Jacobian of $\vec f(\vec x)$. The extremum condition $\vec \nabla s(\vec x^*)=\vec 0$ is equivalent to
\begin{equation} \label{eq:extremum-cond}
    \vec J(\vec x^*)^T \vec f(\vec x^*) = \vec 0 \iff  \vec f(\vec x^*)^T \vec J(\vec x^*) = \vec 0^T. 
\end{equation}
This orthogonality relation, in conjunction with the third condition that $\vec f(\vec x^*) \neq 0$, suggests that the left eigenvectors of the Jacobian at $\vec x^*$ include $\vec f(\vec x^*)$ with a corresponding eigenvalue of zero. 

With the Jacobian in hand, we can now define the \textit{ghost} as a special case of a slow point. Classically, a one-dimensional ghost is a remnant of a saddle-node bifurcation that slows down nearby trajectories \cite{strogatz2018nonlinear} (the term is also used colloquially in higher-dimensional settings \cite{sussillo2013opening}). To generalize this definition to higher dimensions, we first define what it means for a slow point to be transverse stable:
\begin{assumption}[Transverse stability at a slow point]
    Let $\vec x^*$ be a slow point and set $\vec v \coloneqq \vec f(\vec x^*)/\|\vec f(\vec x^*)\|_2$. By Eq. \eqref{eq:extremum-cond}, $\vec v^T \vec J(x^*) = 0$. We assume that for any $\vec w \in  \vec v^\perp\setminus\{\vec 0\}$, defined as collections of $\vec w\neq \vec 0$ satisfying $\vec v^T \vec w =0$, the following statement is true.
\begin{equation}
\label{eq:transverse}
\vec w \in \vec v^\perp\setminus\{0\},\ \vec w^T \vec J(\vec x^*)  = \lambda  \vec w^T \ \implies\ \Re[\lambda] < 0.
\end{equation}
Here $\text{Re}[\lambda]$ refers to the real part of the eigenvalue $\lambda$.
\end{assumption}
Notably, slow points in higher dimensions, including those created after saddle-node bifurcations, can have mixed stability in the transverse directions. However, when all transverse directions are stable, the resulting slow point acts as a \textit{funnel} that rapidly attracts nearby trajectories onto a one-dimensional dynamical bottleneck. As we show in this work, this geometric structure proves particularly valuable for analyzing RNN dynamics. We therefore define ghosts as slow points satisfying this additional constraint:
\begin{definition}[Ghost]
A \emph{ghost} is a slow point $\vec x^*$ that satisfies the transverse stability assumption. 
\end{definition}
As we show next, under the linearized dynamics near a ghost $\vec x^*$, the components transverse to $\vec f(\vec x^*)$ decay exponentially so that local trajectories are rapidly confined to the one-dimensional dynamical bottleneck.

\subsection{Reduction of ghost dynamics to a one-dimensional canonical form} 
With the definition of a ghost point, we have all the pieces ready to perform the reduction to a one-dimensional system. Recall $\vec v = \vec f(\vec x^*) / ||\vec f(\vec x^*)||_2$ such that $\vec v^T \vec v =1$ and let $\vec z(t) = \vec x(t) - \vec x^*$ denote the displacement for some $\vec x(t)$ nearby $\vec x^*$. We perform Taylor expansions of the vector field for small $\vec z(t)$, which we can divide into two components as:
\begin{equation}
    \vec z(t) = \kappa(t) \vec v + \vec z_\perp(t), \quad \vec v^T \vec z_\perp(t) = 0.
\end{equation}
Here, $\kappa(t) = \vec v^T \vec z(t)$ is a scalar parameterizing the evolution along the slow mode when $\vec x(0)=\vec x^*$. Then, to the first order corrections near the ghost point, the flow follows:
\begin{equation}
    \dot {\vec z}(t) = \vec f(\vec x^*+\vec z(t)) \approx \vec f(\vec x^*) + \vec J(\vec x^*) \vec z(t).
\end{equation}
Inserting $\vec z(t)$ into this equation and projecting down to the perpendicular component, we arrive at the equation:
\begin{equation}
\begin{split}
    \dot {\vec z}_\perp(t) &= \vec J(\vec x^*) \vec z_\perp(t) + \kappa(t) \vec J(\vec x^*) \vec v, \\
    &= \vec J(\vec x^*) \vec z_\perp(t) + O(\kappa).
\end{split}
\end{equation}
For $\kappa(0)=0$ and $\vec z_\perp(0) = \vec 0$, we have that $\dot {\vec z}_\perp(0) = \vec 0$ and consequently $\vec z(t\approx0) \approx \vec v\kappa(t)$ with $\dot \kappa(t) = \vec v^T \vec f(\vec x^*+\vec z(t)) \approx ||\vec f(\vec x^*)||_2 \neq  0$. For $t>0$ as long as $\vec z(t)$ remains small, hence Taylor approximation is valid, each element of $\vec z_\perp(t)$ decays rapidly to a value at $\sim O(\kappa)$ since the spectrum of $\vec J(\vec x^*)$ has only negative real parts by the transverse stability assumption. Since we are not interested in the exact flow direction near the ghost, but in the scalar variable $\kappa(t)=\vec v^T \vec z(t)$, we denote the limiting behavior as $\vec z =\vec z_\perp + \kappa \vec v \approx \tilde {\vec v} \kappa$ for some $\tilde {\vec v}$. This allows us to analyze the displacement along $\vec v$ quantitatively by keeping the second-order terms of the Taylor expansion:
\begin{equation}
\begin{split}
    \dot \kappa(t) &= \vec v^T \dot {\vec x}(t) = \vec v^T \vec f(\vec x^*+\vec z) ,\\
    &= ||\vec f(\vec x^*)||_2 +  \frac{\kappa^2(t)}{2} \vec v^T \vec h[\tilde {\vec v},\tilde {\vec v}] + \mathcal{O}(\kappa^3).
\end{split}
\end{equation}
Here, $\vec h[\vec z,\vec z]$ is the vector whose $i$-th entry is $\vec z^T \vec H_i(\vec x^*) \vec z$, with $\vec H_i(x^*) = \vec \nabla^2 f_i(\vec x^*) \in \mathbb R^{N \times N}$. This equation has a constant term $\gamma =||\vec f(\vec x^*)||_2>0$ and a quadratic term with coefficient $ \beta =\frac{1}{2} \vec v^T  \vec h[\tilde {\vec v},\tilde {\vec v}]$ that is assumed to be non-zero. But, there is no linear term since $\vec v^T \vec J(\vec x^*) = 0$ by Eq. \eqref{eq:extremum-cond}. This leads to the reduced dynamics on the slow-mode $\dot \kappa = \gamma  + \beta \kappa^2 + \mathcal{O}(\kappa^3).$ Re-scaling $\kappa \to \kappa/\beta $ and defining $r = \gamma \beta$, we arrive at the equation:
\begin{equation} \label{eq:ghost-prototype}
    \dot \kappa = \kappa^2 + r + O(\kappa^3).
\end{equation}
This simple one-dimensional dynamical system, effectively characterized by a single \textit{scale parameter} $r$ when higher order terms are ignored, describes the universal behavior of the state variables $\kappa(t)$ near the ghost point $\kappa^*=0$. Below, ignoring the $O(\kappa^3)$ corrections, we study the learning dynamics governing the scale parameter $r$ in order to achieve a target passage time of the system variable $\kappa(t)$ through the bottleneck $\kappa=0$. As a side note, since our interest is in speed and not energy, and energy ($\frac{1}{2}\|\vec f(\vec x)\|_2^2$) and speed ($\|\vec f(\vec x)\|_2$) functions both share the same minima, we will refer to $\vec x^*$ as speed minima for the rest of this work.

\subsection{Setting up the canonical model and the delayed-activation task}

\begin{figure}
    \centering
\includegraphics[width=\linewidth]{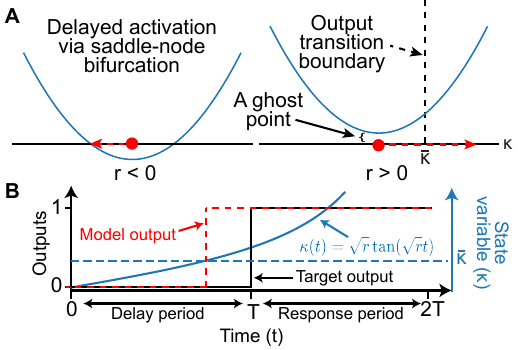}
    \caption{\textbf{Visualization of the canonical model and the delayed-activation (DA) task.} { \textbf{A} In our canonical model, the state variable is initialized at $\kappa(0)=0$ (red dot). \emph{Left.} For $r<0$, the system evolves towards a fixed point at $-\sqrt{-r}$. \emph{Right.} When the system undergoes saddle-node bifurcation as $r$ crosses zero, $\kappa(t)$ diverges (red arrow). A pre-defined threshold, $\bar \kappa$, partitions the model output into two states, with $\kappa>\bar \kappa$ corresponding to an activated response. A ghost point refers to the remnant of a saddle-node bifurcation at the origin, which is a fixed point-like structure emerging from the annihilation of two fixed points. The trajectories considerably slow down and pass the ghost point in times $\propto 1/\sqrt{r}$ \cite{strogatz2018nonlinear}. \textbf{B} A schematic illustration of an example model output for a given $r$.} The DA task requires the model to suppress its output during the delay period and initiate a transition during the response period. In our canonical model, this is achieved by ensuring that the state variable $\kappa(t)$ remains below a predefined threshold $\bar \kappa$ throughout the delay period and crosses it thereafter. In this example, reducing $r$ would be needed to slow down the trajectory, $\kappa(t)=\sqrt{r}\tan(\sqrt{r}t)$, extending the time spent below threshold $\kappa(t)\leq \bar \kappa$.}
    \label{figure1}
\end{figure}

The dynamics near a ghost, as derived in Eq.~\eqref{eq:ghost-prototype}, holds only for small $\kappa$, as higher-order terms can dominate for large $\kappa \gg 1$. However, since our goal is to study the time delay caused by very small parameter values $0 < r \ll 1$, we can focus instead on a simplified canonical model (Fig.~\ref{figure1}\textbf{A}):
\begin{equation} \label{eq:canonical}
    \dot \kappa(t) = r + \kappa^2(t),
\end{equation}
where higher-order corrections have been dropped. Here $\kappa \in \mathbb{R}$ is the scalar state variable and $r \in \mathbb{R}$ is now \emph{learnable}. Since Eq.~\eqref{eq:canonical} has a simple quadratic form, assuming that the initial condition $\kappa(0) = 0$, and for $r>0$, the solution can be written as:
\begin{equation}
\kappa(t) =
\begin{cases}
     \sqrt{r}\,\tan(\sqrt{r}\,t), & \quad t \leq t^*, \\
    \infty, & \quad \text{otherwise,}
\end{cases}
\end{equation}
where $t^* = \tfrac{\pi}{2\sqrt{r}}$ is the finite escape time at which $\kappa(t)$ diverges. This finite-time divergence is not physically meaningful, nor theoretically interesting to our cause, since the higher-order terms in Eq.~\eqref{eq:ghost-prototype} can alter the global landscape, for example by introducing an additional fixed point at $\kappa \gg 0$. 

In this work, we focus on the behavior of $\kappa(t)$ near the origin for small $r \ll 1$. In this parameter regime, the trajectory remains confined near $\kappa \approx 0$ for a duration of $t \sim O(r^{-0.5})$ before escaping the bottleneck \cite{strogatz2018nonlinear,strogatz1989predicted,koch2024ghost}. During this time, since $\kappa(t \sim O(r^{-0.5})) \sim O(\sqrt{r}) \ll 1$, the quadratic term ($\kappa^2 \sim O(r)$) contributes comparably to the constant term ($r$), while higher-order corrections such as $O(\kappa^3) \sim O(r\sqrt{r})$ remain negligible. Once the trajectory escapes this neighborhood and $\kappa(t) \sim O(1)$, higher-order corrections in Eq.~\eqref{eq:ghost-prototype} can have non-negligible effects on the subsequent dynamics. Thus, since our goal is to determine when the system escapes the bottleneck, \textit{i.e.}, the time at which $\kappa(t)$ exceeds a user-defined threshold $\bar\kappa$, Eq.~\eqref{eq:canonical} captures the essential ghost dynamics and provides the appropriate canonical form that we will use in our derivations below.

Returning to the qualitative study of the canonical model, we identify the local fixed point structure: There are two fixed points for $r<0$, and none for $r>0$. The change in the number of fixed points at $r=0$ signals a saddle-node bifurcation, which is well-characterized through bifurcation diagrams \cite{strogatz2018nonlinear}. The slow-down of dynamics near ghost points for $r>0$ is textbook knowledge \cite{strogatz2018nonlinear}, whose computational properties continue to be studied in contemporary work \cite{koch2024ghost}. To study learning dynamics, however, we need a task suited to the innate capabilities of ghosts. Fortunately, ghosts can function as delayed switches \cite{strogatz1989predicted}, in which a state of zero can be assigned to $\kappa \ll O(\sqrt{r})$, and a state of one for $\kappa \gg O(\sqrt{r})$. As we just discussed, when $\kappa$ is initialized near the ghost, this construction achieves a delay of $O(1/\sqrt{r})$ before switching between two states. In line with this observation, we \emph{train} the canonical dynamical system to perform a delayed-activation (DA) task (Fig. \ref{figure1}\textbf{B}). In this task, the state variable is initialized with $\kappa(0) = 0$. We define the model output as $\hat o(\kappa) = \sigma(c(\kappa - \bar \kappa) )$, where $\bar \kappa > 0$ and $c>0$ are pre-defined constants, and $\sigma$ is the sigmoid function. In the limit $c \to \infty$, this output confidently takes on binary values, either $0$ or $1$, hence we call $c$ the \textit{confidence} parameter. 

The objective of the task is for the model's output ($\hat o(\kappa(t))$) to match the target output ($o(t)$) of $0$ until some pre-defined time point $t<T$, and transition to $1$ for $T<t<2T$. Since the only learnable parameter is $r$, the learning process should pick an optimal value, $r^*$, to ensure this condition. Intuitively, one expects the solution to scale as $r^* \sim O(T^{-2})$, so that $\kappa(t)$ escapes the ghost near $\kappa^*=0$ around $t = T$, coinciding with the required switch in output. It is less clear, however, whether minimizing a practical loss function can indeed recover this solution, which is what we investigate next.

\subsection{Analytical derivations of the loss function and its gradient with the canonical model}
We train the canonical model, \textit{i.e.}, update $r$ in Eq. \eqref{eq:canonical}, to perform the DA task by minimizing a loss function:
\begin{equation} \label{eq:loss_original}
    \mathcal{L}(r) = \int_{0}^{2T}  (\hat o(\kappa(t)) - o(t))^2 \diff t.
\end{equation}
For analytical convenience, we assume the limit $c \to \infty$ (referred to as the analytical limit alongside with $\bar \kappa \to \infty$) such that the model output takes the form:
\begin{equation}
    \hat o(\kappa(t)) = \Theta(\kappa(t) - \bar \kappa),
\end{equation}
for some pre-defined $\bar \kappa > 0$, where $\Theta(\cdot)$ is Heaviside function with $\Theta(\kappa) = 1$ if $\kappa>0$, and zero otherwise. Using this output, for $r<0$, the loss function in Eq. \eqref{eq:loss_original} is trivially $T$, since the model output is always $0$. Therefore, we will focus on the case with $r>0$.

In the limit $\bar \kappa \to \infty$, we will simply assume that the network output is $0$ before the escape, \textit{i.e.}, $t \leq t^*$, and $1$ after the escape, \textit{i.e.}, $t \geq t^*$. Using this, we can explicitly compute the loss function:
\begin{equation}
\begin{split}
    \mathcal{L}(r) &= \int_{0}^{2T} \diff t (\hat o(\kappa(t)) - o(t))^2, \\
    &= \int_{0}^{T} \diff t (\hat o(\kappa(t)))^2 + \int_{T}^{2T} \diff t (\hat o(\kappa(t)) - 1)^2.
\end{split}
\end{equation}
There are three important regimes here:
\begin{enumerate}[label = (\roman*)]
    \item $t^* \leq T$,  when the output turns on too early,
    \item $T \leq t^* \leq 2T$, when the output turns on late, and
    \item  $2T \leq t^*$, when the output never turns on during the task.
\end{enumerate}

Let us start with the first case, $t^* \leq T$:
\begin{equation}
\begin{split}
    \mathcal{L}(r) = \int_{t^*}^{T} \diff t  = T-t^* = T - \frac{\pi}{2\sqrt{r}}, \quad \text{for} \quad r\geq \frac{\pi^2}{4T^2}
\end{split}
\end{equation}
Next, we consider the second case, $T \leq t^* \leq 2T$:
\begin{equation}
\begin{split}
    \mathcal{L}(r) = \int_{T}^{t^*} \diff t  = \frac{\pi}{2\sqrt{r}} - T, \quad \text{for} \quad  \frac{\pi^2}{16T^2} \leq r\leq \frac{\pi^2}{4T^2}
\end{split}
\end{equation}

Finally, we consider the case $t^* \geq 2T$:
\begin{equation}
\begin{split}
    \mathcal{L}(r) = T \quad \text{for} \quad   r\leq \frac{\pi^2}{16T^2}.
\end{split}
\end{equation}
When the output never turns on (\textit{i.e.}, $t^* \geq 2T$), the loss function is constant. We will show below that this leads to a failure mode of the model, in which overconfidence in incorrect output (here, $\hat o(t) = 0$ for $t \in [0,2T]$) halts learning. Bringing all cases together, we find the loss function

\begin{figure*}
    \centering
\includegraphics[width=\textwidth]{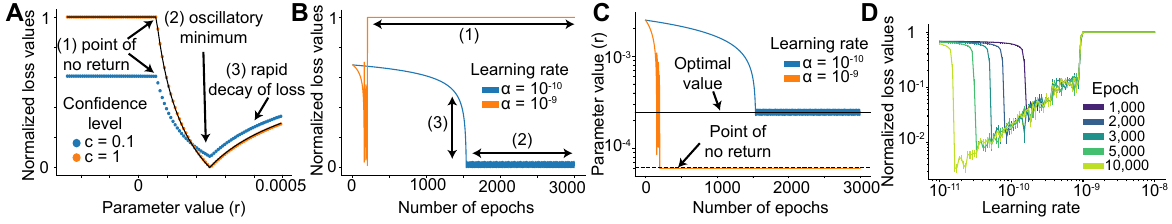} 
    \caption{\textbf{
    The canonical model trained on the DA task exhibits abrupt learning and co-occurring learning instabilities.} \textbf{A} We compared the analytical loss function (black line) vs those computed from realistic parameters (colored dots), in which the model output was defined via a sigmoid function $\hat o(\kappa) = \sigma(c(\kappa-\bar \kappa)).$ The loss function had three distinct regimes: (1) a point of no return, (2) a minimum with non-zero gradient, and (3) rapid decay of the loss function for $r\geq r^*$, where $r^*=\frac{\pi^2}{4T^2}$ is the global minimum. {We illustrate in panel \textbf{B} how each regime affects the training.} Parameters: $T = 100$, $\Delta t = 0.1$, and $\bar \kappa = 10$. \textbf{B-C} Initializing at $r:= 10r^*$, we minimized the loss function values using gradient descent with different learning rates, recapitulating all three regimes in learning dynamics. Plots show loss function (\textbf{B}) and parameter values (\textbf{C}) vs number of epochs. Notably, for $\alpha = 10^{-10}$, even though the loss function decrease abruptly around epoch 1500, the network does not undergo any bifurcations, as evident from $r$ not changing its sign during learning. \textbf{D} Plot shows the loss function values normalized by $T$ at various epochs vs learning rates used for the training process. The canonical model learned best with lower learning rates with more epochs of training. As predicted by Eq. \eqref{eq:critical}, learning is no longer possible for $\alpha \geq 9 * 10^{-10}$. Solid lines: means. Error bars: s.e.m. over 10 training instances, in which $r$ was initialized following a normal distribution that has the mean $10r^*$ and the standard deviation $\frac{r^{*}}{10}$. Parameters for (\textbf{B-D}): $T = 100$ and $\bar \kappa,c \to \infty$ (analytical model).  }
    \label{figure2}
\end{figure*}

\begin{equation} \label{eq:error}
    \mathcal{L}(r) = 
    \begin{cases}
        \left|T-\frac{\pi}{2\sqrt{r}}\right| \quad &\text{for}  \quad r \geq \frac{\pi^2}{16T^2} = \frac{r^*}{4}, \\
        T \quad &\text{otherwise}
    \end{cases}
\end{equation}
and the gradient as:
\begin{equation} \label{eq:gradient}
\nabla \mathcal{L}(r)= 
\begin{cases} 
-\frac{\pi}{4r^{3/2}} & \text{for } \frac{\pi^2}{16T^2} < r < \frac{\pi^2}{4T^2}, \\
\frac{\pi}{4r^{3/2}} & \text{for } r > \frac{\pi^2}{4T^2}, \\
0 & \text{for } r < \frac{\pi^2}{16T^2}.
\end{cases}
\end{equation}
When $r^* = \frac{\pi^2}{4T^2}$, this loss function is zero, achieving the global optimum. {Thus, as expected, the loss function does have a minimum at a value $r^* \sim T^{-2}$.} In this minimum, the state variable ($\kappa(t)$) lingers around the origin with negligible changes up to $T$, before abruptly shooting off to $\infty$. Even though the canonical model has no fixed points for $r^*>0$ (and thus for any $T$), the origin resembles one (for $t \ll T$, $\dot \kappa = r^*+\kappa^2 \approx r^*$ leads to negligible changes in $\kappa(t)\approx 0$), hence the name \textit{ghost}.

\subsection{Extracting actionable insights about abrupt learning with ghost mechanism}

To extract further insights from the analytical loss function and its gradient, we plotted Eq. \eqref{eq:error} and numerically calculated loss function values (with finite $\bar \kappa,c$) with respect to $r$ in Fig. \ref{figure2}\textbf{A}, which revealed three distinct regimes: 
\begin{enumerate}[label = (\arabic*)]
    \item For any $r<r^*/4$, loss minimization would enter a no-learning zone and cannot easily recover due to flat loss function values, \textit{i.e.}, zero gradients ($\nabla \mathcal{L}(r<r^*/4) = 0$) in the analytical limit. We call the boundary of this regime, $r=r^*/4$, as the \textit{point of no return}.
    \item Though $r^*$ is a global minimum, the gradient is discontinuous at $r = r^*$, \textit{i.e.}, $\nabla \mathcal{L}(r\to r^*)$ does not exist. This leads to oscillatory behavior near the minimum, which implies the existence of a critical learning rate ($\alpha^*$) for a naive gradient descent approach (\textit{i.e.}, updates of the form $r \to r - \alpha \nabla \mathcal{L}(r)$, where $\alpha$ is the learning rate), beyond which learning stops. For any $\alpha \geq \alpha^*$, the training process would eventually enter the no-learning zone (where gradients are exactly zero in the limit $c \to \infty$) and can no longer recover.
    \item Picking the learning rate close to $\alpha^*$ is necessary to achieve learning as the loss function quickly saturates to flat values, \textit{i.e.}, $\mathcal L(r) \to T$ for large $r\gg r^*$. Since the loss function also changes abruptly for $\frac{r^*}{4}<r<r^*$ (between $0$ and $T$), minimizing it would lead to an abrupt decline as $r \to r^*$.
\end{enumerate}
{Apart from predicting its existence, our canonical model enables deriving an exact formula for the critical learning rate, $\alpha^*$. Consider the} derivative at $r=r^*$:
\begin{equation}
    \text{At } r = \frac{\pi^2}{4T^2}: \quad \nabla \mathcal{L}(r) = \begin{cases} 
-\frac{2T^3}{\pi^2} & \text{from the left}, \\
\frac{2T^3}{\pi^2} & \text{from the right}.
\end{cases}
\end{equation}
With a learning rate $\alpha^*$, the model would be thrown from the global minimum to the point of no return boundary and get stuck in the no-learning zone during the oscillations if it receives the following update:
\begin{equation} \label{eq:critical}
    \alpha^* \left|\nabla \mathcal{L}(r)\right|_{r \to r^*_+} = \frac{3}{4}r^* = \frac{3\pi^2}{16T^2} \implies \alpha^* = \frac{3\pi^4}{32} T^{-5}.
\end{equation}
Any $\alpha \geq \alpha^*$, the model would eventually pass the point of no return and output only zeros. {Most notably, after rescaling the loss function, and subsequently the gradient, with $1/T$ to match a mean-squared error, the critical learning rate scales with an inverse power law $\alpha^* \sim T^{-4}$, \textit{i.e.}, longer delay timescales make learning progressively harder and require increasingly fine-tuned learning rates.}

To verify the three theoretical insights above, we conducted optimization experiments with naive gradient descent and fixed learning rates (Fig. \ref{figure2}\textbf{B-D}), where a single epoch corresponds to a numerical update with the analytically computed gradient. Starting the training from $r = 10r^*$ and using a learning rate $\alpha < \alpha^*$, we observed an abrupt decay of loss function values around epoch $1500$, followed by oscillations near the global optimum (Fig. \ref{figure2}\textbf{B-C}). In contrast, picking a larger learning rate, $\alpha > \alpha^*$, pushed the network into the no-learning zone (Fig. \ref{figure2}\textbf{B-C}). Moreover, varying the learning rate values confirmed that there was indeed a critical learning rate, $\alpha^*$, beyond which learning did not take place (Fig. \ref{figure2}\textbf{D}). The numerical value of the critical learning rate matched the theoretical prediction ($\alpha^* \approx 9 \ast10^{-10}$) from Eq. \eqref{eq:critical}. Even though smaller learning rates led to lower final errors, learning within few epochs was only achieved with learning rates close to criticality, $\alpha \approx \alpha^*$. 

Finally, since $r$ values are always positive during learning, the abrupt decays in Fig. \ref{figure2}\textbf{B-C} are not preceded by bifurcations, rather through the emergence of ghost points {followed by careful finetuning of the scale parameter $r$.} Interestingly, even though a global minimum with zero loss value exists, the canonical model cannot easily attain it using gradient descent due to complex learning dynamics (Fig. \ref{figure2}), which is not necessarily remedied by using standard optimization strategies, \textit{e.g.}, adding momentum or adaptive methods \cite{kingma2014adam,wilson2017marginal}. In fact, both can exacerbate the problem: momentum accumulates small updates, which can prevent recovery once the system escapes the loss plateau and slightly overshoots $r \sim r^*$; adaptive methods increase the effective step size in flat regions ($r\gg r^*$) by shrinking their gradient history term, which can lead to instability near the parameters that form a ghost ($r\sim r^*$). 

Overall, learning instabilities are expected in over-parametrized RNNs \cite{eisenmann2023bifurcations} or deep learning models solving complex tasks \cite{li2018visualizing}. Here, a simple canonical model with just one state variable and one parameter already captures several complex learning behaviors commonly observed in practice, such as oscillatory minima (see blue lines in Fig. \ref{figure2}\textbf{B-C}) and a range of effective learning rates (Fig. \ref{figure2}\textbf{D}).

\section{Ghost dynamics in low-rank RNNs} \label{sec:lowrank}

To test whether ghosts emerge during training in RNNs, and whether their formation follows the predictions of the canonical model, we study the learning dynamics in low-rank RNNs trained to solve working memory tasks, which we show involve formation of ghosts. Low-rank RNNs provide a natural testbed for this investigation. On the one hand, their dynamics can be reverse-engineered and interpreted through low-dimensional phase portraits \cite{valente2022extracting,dubreuil2022role,dinc2025latent}, offering direct access to the mechanisms that emerge during training, \textit{i.e.}, an advantage not available in full-rank counterparts (see Section \ref{sec:results}). On the other hand, focusing on low-rank dynamics is not overly restrictive: even full-rank RNNs often converge to low-dimensional solutions \cite{valente2022extracting}, and our canonical model already captures local one-dimensional dynamics near ghost points (Section~\ref{sec:theory}). For these reasons, we begin with rank-one and rank-two RNNs, where interpretability is clearest and direct links to the canonical model can be drawn, whereas Section~\ref{sec:results} will generalize the insights derived from low-rank RNNs to the case of full-rank RNNs.

Specifically, we analyze a representative class of RNNs with the following equations:
\begin{equation} \label{eq:rnn_eq}
\tau \dot {\vec x}(t) = -\vec x(t) + \tanh(\vec W \vec x(t) + \vec W^{\rm in} \vec u(t) + \vec b +\vec \epsilon),
\end{equation}
where $\vec x(t) \in \mathbb R^N$ are firing rates of $N$ neurons, $\tau>0$ is the neuronal decay time, $\vec W \in \mathbb R^{N\times N}$ and $\vec b \in \mathbb R^N$ are trainable weights and biases. Here, $\vec u(t)\in \mathbb R^{N_{\rm in}}$ refers to optional control inputs, which we will use for the working memory tasks below, and $\vec W^{\rm in} \in \mathbb R^{N \times N_{\rm in}}$ corresponding input weights, and $\vec \epsilon$ is an optional noise term that is set to zero unless explicitly stated. In this section, we focus on rank-one or rank-two RNNs, in which $\vec W$ is constrained to have $N-1$ or $N-2$ zero singular values, respectively.

\subsection{Connecting ghost mechanism to rank-one RNNs}
Rank-one RNNs are theoretically capable of solving the DA task by virtue of their universal approximation property \cite{beiran2021shaping,dinc2025latent}, \textit{i.e.}, they can approximate the canonical model in Eq. \eqref{eq:canonical} with arbitrarily small error. To make this connection explicit, we first enforce a rank-one structure and show that the resulting high-dimensional dynamics reduce to a one-dimensional latent dynamical system. This reduction of Eq. \eqref{eq:rnn_eq} is standard \cite{dinc2025latent} and has also been established in alternative RNN formulations \cite{mastrogiuseppe2018linking,dubreuil2022role,beiran2021shaping}. Building on this reduction, we then show that the latent system’s parameters can be mapped directly onto the scale parameter $r$ of the canonical model. This mapping links the abstract, analytical ghost dynamics to the concrete training dynamics in low-rank RNNs and provides a principled account of nonlinear learning dynamics, which we will illustrate in the next subsection.

We start by enforcing the rank constraint, \textit{i.e.}, by constructing $\vec W = \sum_{k=1}^K \vec m^{(k)} \vec n^{(k)T}$ for $\vec m^{(k)} \in \mathbb R^N$ and $\vec n^{(k)} \in \mathbb R^N$ and $k=1,\ldots,K$ denoting the rank. Using this relationship and assuming $K=1$ and no inputs for simplicity (we now drop the superscript $k$; derivations for larger ranks $K>1$ follow similarly), we can define a latent variable, $\kappa(t) = \vec n^T \vec x(t)$, such that the RNN equations transform into a one-dimensional form:
\begin{equation} \label{eq:latent_eq}
    \tau \dot \kappa(t) = - \kappa(t) + \vec n^T \tanh(\vec m \kappa(t) + \vec b) = f(\kappa; \vec m,\vec n,\vec b),
\end{equation}
which we refer to as the latent dynamical system {\cite{dinc2025latent}}. In this transformed picture, training the RNN will implicitly update the parameters of the low-dimensional dynamical system, leading to the learning of desired dynamics ($f(\kappa; \vec m,\vec n,\vec b)$) \cite{dinc2025latent,beiran2021shaping}. We do not impose constraints on the latent system itself. Instead, the RNN is optimized to solve the task with a readout defined from the latent variable, $\hat o(t) = \psi(\kappa(t))$, where $\psi(\cdot)$ is a parametrized function (e.g., a sigmoid; see below). This procedure allows us to use the learned parameters ($\vec m, \vec n, \vec b$) and Eq.~\eqref{eq:latent_eq} to visualize the latent dynamical system that emerges during training. (For further discussions of latent dynamical systems in low-rank RNNs satisfying Eq. \eqref{eq:rnn_eq}, see Ref. \cite{dinc2025latent}.)

For now, we assume training has converged on a task requiring ghosts, so that a local speed minimum has emerged at $\kappa = \kappa^*$. Then, we can write the learned flow near $\kappa^*$ as:
\begin{equation}
\tau \dot \kappa = f(\kappa) = \gamma + \beta(\kappa - \kappa^*)^2 + O((\kappa-\kappa^*)^3),
\end{equation}
where we recall the connection $r = \gamma \beta$ from Eq. \eqref{eq:ghost-prototype} and denote {$f(\kappa;\vec m,\vec n, \vec b)$} concisely as $f(\kappa)$. Our goal is to connect $r$ to the set of parameters $(\vec m,\vec n,\vec b)$. We first note that the speed minimum at $\kappa = \kappa^*$ leads to the condition:
\begin{equation}
    f'(\kappa^*) = 0 \implies  \sum_{i=1}^N n_i m_i (1-\tanh^2(m_i\kappa^*+b_i)) =1.
\end{equation}
The corresponding values of $\gamma$ and $\beta$ in terms of $(\vec m,\vec n,\vec b)$ are:
\begin{subequations}
    \begin{align}
        \gamma & = -\kappa^* + \sum_{i=1}^N n_i \tanh(m_i \kappa^* + b_i), \\
        \beta & = - \sum_{i=1}^N n_i m_i^2 \tanh(m_i \kappa^* + b_i) (1-\tanh^2(m_i \kappa^* + b_i))
    \end{align}
\end{subequations}
Notably,
\begin{equation} \label{eq:eff_r}
r = \beta \gamma 
= g(\vec m,\vec n,\vec b,\kappa^*(\vec m,\vec n,\vec b)),
\end{equation}
constitutes a function that depends on the parameters $(\vec m,\vec n,\vec b)$ explicitly, while  $\kappa^*$ depends on these parameters implicitly through the constraint. This function is too complicated to admit a closed-form analytical expression, at least within the knowledge of the authors.  Nevertheless, for finite $N$, all components are analytic, involving only algebraic operations  and the smooth nonlinearity $\tanh(\cdot)$. Consequently, the change in $r:=r(\vec m, \vec n, \vec b)$ under infinitesimal perturbations $(\vec \Delta m,\vec \Delta n,\vec \Delta b)$ is linear. Formally, by Taylor expansion we obtain
\begin{equation}\label{eq:taylor}
\begin{split}
&r(m_1+\Delta m_1,\ldots,n_1+\Delta n_1,\ldots,b_1+\Delta b_1,\ldots) \\
&= r(m_1,\ldots,n_1,\ldots,b_1,\ldots) 
+ \Delta r + \mathcal{O}(\|(\vec \Delta m,\vec \Delta n,\vec \Delta b)\|_2^2),
\end{split}
\end{equation}
with the first-order increment
\begin{equation}\label{eq:dr}
\begin{split}
\Delta r 
&= \sum_{i=1}^N \frac{\partial r}{\partial m_i}\,\Delta m_i
+ \sum_{i=1}^N \frac{\partial r}{\partial n_i}\,\Delta n_i \\
&\quad+ \sum_{i=1}^N \frac{\partial r}{\partial b_i}\,\Delta b_i.
\end{split}
\end{equation}
Each partial derivative contains both explicit contributions from $\vec m,\vec n,\vec b$, and implicit contributions through $\kappa^*(\vec m,\vec n,\vec b)$. Thus, for sufficiently small parameter changes $(\vec \Delta m,\vec \Delta n, \vec \Delta b)$, \textit{e.g.}, under a small learning rate, Eq.~\eqref{eq:dr} describes how parameter updates induce a change in $r$. Thus, a gradient descent on $(\vec{m}, \vec{n}, \vec{b})$ would change the parameters of the low-rank RNN, but can we connect the resulting change to the gradient-driven parameter updates of the ghost mechanism?

Fortunately, Eq.~\eqref{eq:dr} provides the necessary link between $\Delta r$ and $\nabla \mathcal{L}(r)$. To see why, note that the one-dimensional reduction near the ghost is assumed to capture the observed neural activities during task performance. Since the loss is computed from these activities, we assume that perturbations of $(\vec{m}, \vec{n}, \vec{b})$ that leave $r$ invariant leave the loss approximately unchanged, so that $\mathcal{L}(\vec{m}, \vec{n}, \vec{b}) \approx \mathcal{L}(r)$. Now, we recall that the canonical model prescribes this change directly as
\begin{equation}\label{eq:r-update}
\Delta r = -\alpha \nabla \mathcal{L}(r),
\end{equation}
with learning rate $\alpha$. In the low-rank RNN, by contrast, $r$ is not updated directly but through parameter updates,
\begin{subequations}\label{eq:rnnupdates}
\begin{align}
\Delta m_i &= -\alpha \,\frac{\partial \mathcal{L}(\vec m,\vec n,\vec b)}{\partial m_i}, \\
\Delta n_i &= -\alpha \,\frac{\partial \mathcal{L}(\vec m,\vec n,\vec b)}{\partial n_i}, \\
\Delta b_i &= -\alpha \,\frac{\partial \mathcal{L}(\vec m,\vec n,\vec b)}{\partial b_i},
\end{align}
\end{subequations}
for $i=1,\ldots,N$. Substituting Eq.~\eqref{eq:rnnupdates} into Eq.~\eqref{eq:dr} connects the gradient dynamics in Eq.~\eqref{eq:r-update} to the learning dynamics of low-rank RNN by
\begin{equation}
\nabla \mathcal{L}(r)
\approx \sum_{\theta \in (m_i,n_i,b_i)} 
h_\theta(\vec m,\vec n,\vec b)\,\frac{\partial \mathcal{L}(\vec m,\vec n,\vec b)}{\partial \theta},
\end{equation}
for coefficients $\vec h(\vec m,\vec n,\vec b) = \vec \nabla_\theta r $ (where $\vec \nabla_\theta$ refers to the gradient with respect to the parameters $(\vec m,\vec n, \vec b)$) that depend on the parameters forming the ghost in the first place. This equation is simply the inner product of the coefficient vector $\vec h(\vec m,\vec n,\vec b)$ and the gradient with respect to RNN parameters $\vec \nabla_\theta \mathcal{L}(\vec m,\vec n,\vec b)$. Thus, the update to the canonical parameter $r$ can be written as a projection of the high-dimensional gradient of low-rank RNN parameters:
\begin{equation} \label{eq:connection}
\begin{split}
\Delta r &= - \alpha {\nabla  \mathcal{L}(r)}, \\
&\approx - \alpha \vec h(\vec m,\vec n,\vec b)^T \vec \nabla_\theta \mathcal{L}(\vec m,\vec n,\vec b), \\
&= \vec h(\vec m,\vec n,\vec b)^T \vec \Delta \theta,
\end{split}
\end{equation}
{where $\vec \Delta \theta = -\alpha \vec \nabla_\theta \mathcal{L}(\vec m,\vec n,\vec b)$ refers to the parameter update in the rank-one RNN.} This establishes a direct linear connection between the update of the canonical parameter $\Delta r$ and the update of the low-rank RNN parameters $\vec \Delta \theta$: the former is obtained as a projection of the latter onto $\vec \nabla_\theta r$. 

In essence, Eq.~\eqref{eq:connection} is an application of the chain rule to the dependency $r:=r(\vec m,\vec n,\vec b)$, relying on the analyticity of this mapping and holding under the conditions of small learning rate (which is necessary for learning a large timescale; as derived in Section~\ref{sec:theory}) and correspondingly small parameter updates. Under these assumptions, vanishing slopes, ill-conditioned directions, or bifurcations that hinder learning in the canonical dynamics are likewise expected to appear in the learning dynamics of the low-rank RNN parameters as we show next.

\begin{figure*}
    \centering
\includegraphics[width=\textwidth]{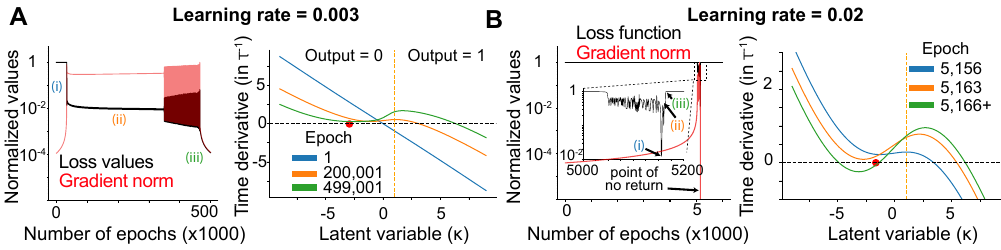}
    \caption{\textbf{A rank-one RNN trained on the DA task reproduced the main findings of the canonical model.} We trained a rank-one RNN on the DA task, in which the output of the RNN was defined as $\hat o(\kappa) = \sigma(c(\kappa - 1))$. Here, $\kappa$ is the latent variable and $c$ is the confidence level. \textbf{A} {\textit{Left.}} The RNN trained with a relatively low learning rate showed the abrupt jump in the loss function (between (i) and (ii)), and had oscillatory behavior before converging to a minimum (between (ii) and (iii)). {\textit{Right.}} The resulting network learned local ghost points with a small, but non-zero, distance from the $y=0$ line. \textbf{B} When training the same network with higher learning rate, a saddle-node bifurcation occurred, putting the network beyond the point of no return. The network could no longer recover, as indicated by the practically zero gradient after the bifurcation. Parameters: $\tau = 10ms$, $\Delta t = 5ms$, $T = 100ms$, $N=100$ neurons, $c=10$. {We sampled initial RNN parameters following $m_i \sim \mathcal{N}(0,N^{-1})$, $n_i \sim  \mathcal{N}(0,N^{-1})$, and $b_i = 0$ for $i=1,\ldots,N$ neurons. Here, $\mathcal{N}(\mu,\sigma^2)$ refers to a normal distribution with mean $\mu$ and variance $\sigma^2$.} We initialized all units to be $x_i(0) = -0.3$ and used stochastic gradient descent. Red dots correspond to the initial values of $\kappa(t)$ for the final networks. ``Gradient norm'' refers to the $\mathcal L_2$ norm of the (flattened) gradient computed over all trainable parameters. Training videos for multiple seeds are available via the Zenodo link in the data availability statement.}
    \label{figure3}
\end{figure*}

\subsection{Reproducing theoretical results from the canonical model}

We now turn to the learning dynamics of rank-one RNNs trained to solve the DA task and compare them to the predictions of the canonical model. For this task, we define the RNN output as
\begin{equation} \label{eq:lat-red}
\hat o(t) = \sigma(c (\kappa - \bar\kappa)),
\end{equation}
for some $\bar \kappa>0$. Training the network corresponds to minimizing the difference between $o(t)$ and $\hat o(t)$ while updating the parameters $(\vec m,\vec n,\vec b)$. Unless otherwise stated, for simplicity, we fix $\bar \kappa = 1$ (without loss of generality) and $c = 10$, though we will also report training results with learnable $c$ below.

As a first step, we confirmed that rank-one RNNs exhibit the ghost mechanism to solve the DA task when trained with a suitable learning rate (Fig. \ref{figure3}\textbf{A}). During learning, we observed the emergence of a (local) ghost point (Fig. \ref{figure3}\textbf{A}), which {coincided with sharp transitions} in both the loss function (abrupt decay) and the $L_2$ norm of the gradient (abrupt increase). Importantly, the latent variable near the ghost point had very slow time-dynamics, leading to slow changes in latent activations. By finetuning the distance at the ghost point (similar to $r$ in Eq. \eqref{eq:canonical}), RNN learns to wait for a delay period of $T$ and output one only after the latent variable escapes the ghost point. When we studied the development of this ghost point in the latent dynamical system, we uncovered that the latent dynamical system has always had one fixed point (there was only one fixed point after every epoch within $\kappa \in [-15,15]$). Consequently, as in our canonical model, this RNN created the ghost point without undergoing any bifurcations, {but by carefully fine-tuning the minimum speed value at the ghost.}

Similar to the canonical model, the learning dynamics in rank-one RNNs also showed the oscillatory behavior for low loss function values (Fig. \ref{figure3}\textbf{A}). Yet, unlike the canonical model, we observed that the rank-one RNN was able to eventually escape the oscillations and decrease the loss further. Studying the learned latent dynamical system revealed that the escape was accompanied with a sharpening of the curvature around the ghost (Fig. \ref{figure3}\textbf{A}), whereas our canonical model had a fixed curvature regardless of $r$. With the increased curvature, $\kappa(t)$ spent less time in uncertain states around $\bar \kappa$, {\textit{i.e.}, those with $\hat o \approx 0.5$, and the output abruptly transitioned from zero to one.} Though prior work has shown that networks with many parameters are more expressive compared to their counterparts \cite{beiran2021shaping}, this observation suggests an added benefit: They may have desirable loss landscapes to achieve the global minima (which may exist but remain unattainable) in less expressive models, echoing earlier work on Gaussian random fields in high-dimensional spaces \cite{bray2007statistics}.

Next, { we increased the learning rate to test the prediction of criticality. As predicted, this prevented learning altogether} (Fig. \ref{figure3}\textbf{B}). Specifically, we examined the latent dynamical system evolution in a rank-one RNN trained with a high learning rate (Fig. \ref{figure3}\textbf{B}). Similar to the canonical model, the network underwent a saddle-node bifurcation and remained stuck in a no-learning zone with a nearly zero gradient. Though one might think that the fixed nature of the output confidence was the reason, we observed that another network, in which $c$ was trainable, still got stuck in the no-learning zone (see the data and code available via Zenodo link shared at the end). 

{
In summary, low-rank RNNs, whose high-dimensional neural activities are described by a one-dimensional latent dynamical system, displayed all three properties predicted by the canonical model: abrupt loss decreases with ghost formation, oscillatory minima near the optimum, and a critical learning rate that can drive the network into a no-learning zone.}

\begin{figure*}
    \includegraphics[width=\linewidth]{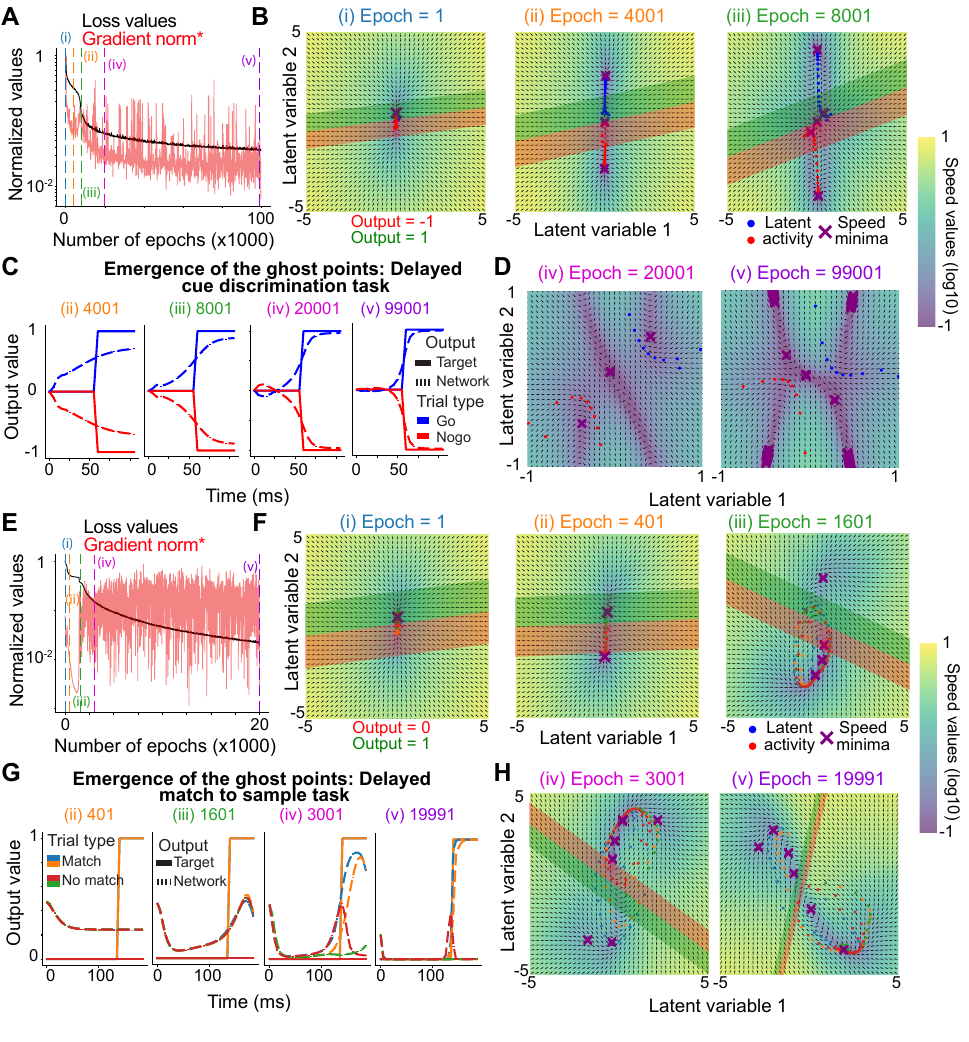}
    \vspace{-3em}
    \caption{\textbf{Rank-two RNNs trained on two commonly studied working memory tasks exhibited ghost mechanisms.} \textbf{A-D} A rank-two RNN was trained on a delayed cue discrimination task, where the network observed one of two cues ($+1$ for Go, $-1$ for NoGo) and, after a short delay, produced an output indicating the cue identity. \textbf{A} Training dynamics were characterized by abrupt transitions, oscillatory gradients, and sharp spikes in the loss values. {Five representative epochs were marked with colored labels and discussed in subsequent panels.} Data points for every $100$ epochs are shown. “Gradient norm$^*$” denotes the sum of Frobenius norms of gradients for the low-rank components of the recurrent weight matrix.  \textbf{B} Latent phase portraits revealed distinct mechanisms underlying these transitions. Between (i) and (ii), the network learned a bistable dynamical system, with attractors corresponding to Go and NoGo output states. Between (ii) and (iii), the RNN learned two trial-type-specific ghost points, prolonging latent trajectories near the decision boundary. Flow speed is color-coded, arrows show normalized flow directions, red/green bands denote the decision band ($\hat o \in [-0.5,0.5]$ for a $\tanh(\cdot)$ output nonlinearity and $\hat o \in [0.27,0.73]$ for a $\sigma(\cdot)$ output nonlinearity), purple crosses mark speed minima {(identified numerically from the phase portraits with nearest neighbor comparisons)}, and blue/red circles track latent states for representative Go/NoGo trials.  \textbf{C} {Plots show network outputs across distinct epochs. Dashed lines denote averages within trials of the same type; the s.e.m. values are negligible.} \textbf{D} A close-up of the latent phase portrait at two later epochs, marked with (iv) and (v){, in which ghost timescales are finetuned and extended slow regions emerge, respectively.}  {(Caption continues in the next page)}}  \label{figure4}
\end{figure*}

\begin{figure*}
\justifying
\noindent
{FIG.~\thefigure\ (continued). \textbf{E-H} Same as in \textbf{A-D}, but for a rank-two RNN trained on a delayed match to sample task, in which the network receives two  input cues ($\pm1$) separated by a delay and has to output whether these cues match (1) or not (0). Similar to before, RNN underwent several regimes of learning (denoted as (i-iv) in (\textbf{E}), with each mechanism (\textbf{F}) endowing the network output with a unique capability (\textbf{G}). In the final solution, ghosts emerged around and on an oscillatory cycle to delay the network output and to aid matching of the two cues (\textbf{H}). Parameters: $\tau=10$ ms, $\Delta t=5$ ms, $N=100$ neurons. Initial neural activity values (batch size of $500$ for each trial type) were sampled as {$x_i(0) \sim \tanh(\mathcal{N}(0,0.3^2))$. Recurrent weights for the rank-2 networks are defined as $\vec W = \vec M \vec N$, with initializations $M_{ij} \sim \mathcal{N}(0,1)$, $N_{ji}\sim \mathcal{N}(0,N^{-1})$ for $i=1,\ldots,N$ and $j=1,2$, such that $\vec W$ is constrained to be rank-2. Input and output weights are initialized using Xavier uniform initialization. Neuronal biases are initialized via $b_i \sim \mathcal{N}(0,0.1^2)$ for $i=1,\ldots,N$ and the output bias is initialized by sampling uniformly from $[-1/\sqrt{2},1/\sqrt{2}]$.} For \textbf{A-D}, the network was trained with gradient descent (learning rate $10^{-3}$); $T_{\rm inp}=10$ms, $T_{\rm delay}=50$ms, and $T_{\rm resp}=50$ms. No additional noise was added to the neurons during forward evolution. For \textbf{E-H}, the network was trained with ADAM \cite{kingma2014adam} (learning rate $10^{-4}$); $T_{\rm inp}=30$ms {for both input windows}, $T_{\rm delay}=80$ ms, and $T_{\rm resp}=50$ms. A random noise with zero mean and $10^{-2}$ s.d. was added to individual neural activities at every time step. Training videos for multiple seeds are available via the Zenodo link in the data availability statement. }
\end{figure*}

{
\subsection{Ghost mechanism in working memory tasks}

Having established the ghost mechanism in the simple delayed-activation task, we next asked whether similar dynamics also emerge in more demanding working-memory paradigms \cite{masse2019circuit,yang2019task}. We focused on two canonical tasks: delayed cue discrimination (DCD) and delayed match-to-sample (DMTS). In both tasks, the RNN receives an input cue of value $\pm 1$ for a duration $T_{\rm inp}$, followed by a delay interval $T_{\rm delay}$ during which the network must retain this information. In the DCD task, after the delay, the RNN is required to output $\pm 1$ corresponding to the original cue for a response window $T_{\rm resp}$. In the DMTS task, a second cue is presented after the delay for another $T_{\rm inp}$ interval, and the RNN must then report whether the two cues match by producing $1$ (match) or $0$ (non-match) during the response window $T_{\rm resp}$.

Although rank-one RNNs might in principle solve these tasks, in practice they failed to converge to accurate solutions. We therefore considered rank-two RNNs, whose activity {(assuming no noise)} is effectively captured by a two-dimensional latent dynamical system{, which was at times driven by some control inputs $\vec u(t) \in \mathbb R^2$ depending on the task}:
\begin{equation}
\begin{split}
     \tau \dot \kappa_j(t) &= -\kappa_j(t) + \vec n^{(j)T}\tanh\left({ \vec z(t)}\right), \\
    { \vec z(t)} &{= \sum_{i=1}^2 \vec m^{(i)}\kappa_i(t) + \vec W^{\rm in} \vec u(t)+ \vec b}
\end{split}
\end{equation}
where the latent variables are defined as {$\kappa_j(t) = \vec n^{(j)T}\vec x(t)$ for $j=1,2$}. {The output is readout from latent variables via $\hat o(t) = \vec w_{{\rm out}}^T \vec \kappa(t) + b_{\rm out}$, where $\vec w_{\rm out} \in \mathbb R^2$ and $b_{\rm out} \in \mathbb R$ are output weight and bias parameters.} Training the RNN parameters therefore corresponds to updating the parameters of this low-dimensional system, just as in the one-dimensional setting of Eq.~\eqref{eq:latent_eq}. We used this formulation to visualize and study the learned latent dynamics via two-dimensional phase portraits {in the absence of control inputs} (Fig.~\ref{figure4}).

In both tasks, training trajectories exhibited abrupt loss drops proceeded by oscillatory gradient norms, coinciding with the emergence and subsequent tuning of the ghost mechanism (Fig.~\ref{figure4}). For the DCD task, two sharp loss transitions appeared within the first $8{,}000$ epochs (Fig.~\ref{figure4}\textbf{A}, markers (i)-(iii)). Examining the latent phase portraits at these transition points revealed distinct mechanisms underlying each jump (Fig.~\ref{figure4}\textbf{B-D}). During the first jump, between (i) and (ii), the RNN underwent a bifurcation and acquired bistable dynamics with attractors corresponding to the Go and NoGo outputs (Fig.~\ref{figure4}\textbf{B}, \textit{middle}). These new fixed points enabled discrimination of the input cues, but because no mechanism yet existed to delay the output, the RNN produced premature responses (Fig.~\ref{figure4}\textbf{C}, \textit{first panel}). The second jump in the loss, between (ii) and (iii), led to the formation of two cue-specific ghosts (Fig.~\ref{figure4}\textbf{B}, \textit{right}). These bottlenecks prolonged trajectories in a cue-dependent manner, suppressing premature outputs during the delay and enabling the timed transition during the response window (Fig.~\ref{figure4}\textbf{C}, \textit{second panel}). In later epochs, marked as (iv) and (v) in Fig.~\ref{figure4}\textbf{A}, the loss decreased gradually {as the timescales near the ghosts were fine-tuned (Fig.~\ref{figure4}\textbf{D}, \textit{left}), and eventually extended slow regions formed (Fig.~\ref{figure4}\textbf{D}, \textit{right}). Ultimately, these changes allowed} the RNN to delay its output for precisely the required duration (Fig.~\ref{figure4}\textbf{C}, \textit{last two panels}).

A similar sequence of learning processes unfolded in the learning of the DMTS task (Fig.~\ref{figure4}\textbf{E-H}). In the early epochs, marked as (i)-(iii) in Fig.~\ref{figure4}\textbf{E}, training again exhibited abrupt loss changes followed later by oscillatory gradient norms. At this stage, the latent dynamics failed to reliably separate match from mismatch trajectories (Fig.~\ref{figure4}\textbf{F-G}). While several transient ghosts emerged in this window (Fig.~\ref{figure4}\textbf{F}), these were not yet arranged to maintain or compare cue information across or after the delay, respectively. As a result, network outputs resembled the behavior of the simpler delayed-activation task, responding in the same way regardless of the input patterns rather than performing the intended cue comparison (Fig.~\ref{figure4}\textbf{G}, \textit{first two panels}).

{In later epochs}, the network had acquired the ability to correctly distinguish and compare the cues (Fig.~\ref{figure4}\textbf{G}, {\textit{last two panels}}). At this stage, several oscillatory cycles had formed in the latent phase portraits, which enabled cue comparison, with ghosts positioned in their vicinity to slow down the dynamics as needed (Fig.~\ref{figure4}\textbf{H}). This configuration gated progression within the latent dynamics, forcing trajectories to linger near {a ghost} until the second cue arrived, \textit{i.e.}, either knocking them out to a larger cycle that produced a positive output, or leaving them confined to smaller cycles that yielded zero outputs (Fig.~\ref{figure4}\textbf{H}, \textit{right}). In this way, the network preserved information about the initial cue during the delay and generated a properly timed match vs. no-match response after the second stimulus (Fig.~\ref{figure4}\textbf{G}, \textit{final panel}).

Taken together, our results on the DCD and DMTS tasks suggest that rank-two RNNs can acquire working-memory behavior through transient, ghost-induced bottlenecks in a two-dimensional latent space. These bottlenecks emerged alongside loss jumps and enabled short-lived memory via slow passage near ghosts, providing an effective delay without trapping the dynamics in stable attractors. Thus, although our study of the delayed-activation task may appear to be a highly constrained scenario, ghosts arise naturally across distinct working-memory tasks.

}

\section{Insights for full-rank RNNs} \label{sec:results}

\begin{figure*}
    \centering
    \includegraphics[width=\textwidth]{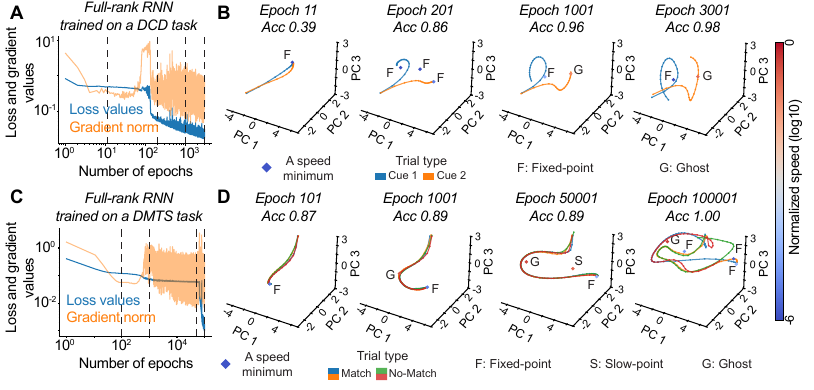}
    \caption{\textbf{Representative full-rank RNNs exhibited ghost mechanisms when trained on working memory tasks.} To illustrate that full-rank RNNs can learn ghost mechanism to support working memory, we studied a full-rank RNN trained on a DCD task (\textbf{A-B}) and another one trained on a DMTS task (\textbf{C-D}). \textbf{A} The learning progression curves as a function of epochs. Gradient norm refers to the Frobenius norm of the gradient corresponding to the recurrent weight matrix. Most significant jumps occur within the first {$\sim 100-200$}, with subsequent epochs refining the learning solution and gradually decreasing the loss function values. \textbf{B} We extracted the speed minima from the neural activities of full-rank RNNs and studied their progression over learning epochs. The plots show the neural trajectories projected down to the top three principal components. Diamonds belong to speed minima found by initializing our numerical solver with the samples along the trajectories. The two speed minima at {epochs 1001 and 3001} (with orange colors) are ghosts, characterized by each Jacobian having one effectively zero and otherwise all negative eigenvalues. Others are fixed points (though all are not necessarily attractive). {The color scale on the right denotes the normalized speed ($||\tau \dot {\vec x}||_2$) of the fixed, slow, or ghost points.} \textbf{C-D} Same as in panels \textbf{A-B}, but for an RNN performing a DMTS task. Notably, this RNN also utilized a slow point during learning transitions, which is distinct from ghosts in the sense that its Jacobian can have positive eigenvalues (compare panels \textbf{B} and \textbf{D}). Parameters:  $\tau = 10ms$, $\Delta t = 5ms$, $T_{\rm inp}= 10ms$, $T_{\rm delay}=50ms/150ms$ for panels \textbf{(B)}/\textbf{(D)}, respectively, $T_{\rm resp}=50ms$, and $N=100$ neurons. {Firing rates are initially sampled following $x_i(0) \sim \tanh(\mathcal{N}(0,0.1^2))$}, iid noise with an s.d. of $10^{-2}$ is added to each neuron during every step of the time evolution. {We initialized the recurrent, input, and output weights using Xavier uniform initialization \cite{glorot2010understanding}. Neuronal biases are initialized from the uniform distribution on $[-1,1]$, whereas the output bias is taken from the uniform distribution on $[-0.1,0.1]$.} We trained the networks with gradient descent, each batch containing a single trial from each type and $\alpha = 10^{-2}$. {Accuracies (Acc.) denote the average agreement between ternarized (thresholds $\pm 0.33$; \textbf{A-B}) or binarized (threshold $0.5$; \textbf{C-D}) network outputs and target outputs across all times and trials.}}
    \label{figure5}
\end{figure*}

We now extend our analysis to full-rank RNNs, where lifting the rank restriction expands both the parameter space and the dimensionality of the underlying dynamical system, enabling the networks to explore higher-dimensional manifolds and exhibit qualitatively richer behaviors \citep{strogatz2018nonlinear}. In this section, we first illustrate that ghosts can emerge in full-rank RNNs during learning. Then, we continue by studying three key aspects that reveal both shared principles and distinctive features between low- and full-rank RNNs. First, both exhibit critical learning rates beyond which learning fails, though full-rank networks reach criticality at higher values and display more stable dynamics overall. Second, oscillatory gradients and associated no-learning zones emerge in both cases, demonstrating the generality of these phenomena. Third, we demonstrate that these no-learning zones arise from the properties of the learned dynamical system and its outputs, rather than architectural assumptions. By designing an output mapping that controls confidence levels, we show that networks can escape no-learning zones and resume learning, validating our mechanistic understanding.

\subsection{Ghost dynamics in full-rank RNNs}

We now turn to full-rank RNNs by lifting the rank restriction on the weight matrix $\vec W \in \mathbb R^{N\times N}$ {and study full-rank networks whose dynamics are described by Eq. \eqref{eq:rnn_eq}}. Prior studies have shown that such networks often develop speed minima that are not fixed points, known as \textit{slow point manifolds}, when trained on working-memory tasks \cite{sussillo2013opening,haputhanthri2024why,maheswaranathan2019reverse,ribeiro2020beyond,maheswaranathan2019universality}. We now show that full-rank RNNs develop the same ghost bottlenecks introduced in Section~\ref{sec:theory}\textbf{A} and demonstrated in low-rank networks (Figs.~\ref{figure3}, \ref{figure4}). These bottlenecks constitute a special form of slow points that funnel nearby trajectories through transient slow regions in state space and can be implemented without explicit dimensionality restrictions.

\begin{figure*}
    \centering
    \includegraphics[width=\textwidth]{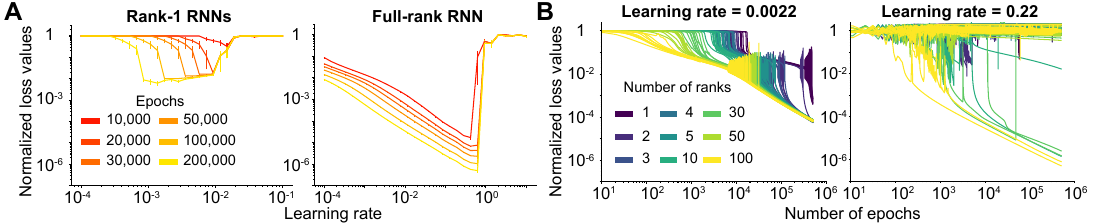}
    \caption{\textbf{Increasing the ranks of the RNNs available for training partially stabilized learning dynamics in the DA task.} \textbf{A} We performed the analysis in Fig. \ref{figure2}\textbf{D} for the rank-one (\emph{left}) and full-rank (\emph{right}) RNNs trained on the DA task. We reported means and standard errors over ten random seeds. \textbf{B} We trained RNNs with varying ranks (1, 2, 3, 4, 5, 10, 30, 50, 100) on the DA task. Each line corresponds to a run with a different seed. \emph{Left.} The RNN trained with a relatively low learning rate showed the abrupt jump in the loss function values and the following oscillations, similar to Fig. \ref{figure3}\textbf{A}. Increasing the number of trainable ranks shortened the learning process, with milder and sustained decreases in the loss function. \emph{Right.} For a substantially high learning rate, {few RNNs that could train without entering the no-learning zone had higher ranks (here, $K\geq 10$). This contrasts with the $K=1$ case that is sufficient to solve the task. See the Zenodo repository for additional data involving varying learning rates, where lower-rank networks progressively fail to train with increasing $\alpha$.} Parameters: $\tau = 10ms$, $\Delta t = 5ms$, $T = 100ms$, and $N=100$ neurons. We initialized all units to be $x_i(0) = -0.3$ {for $i=1,\ldots,N$} and used stochastic gradient descent. For rank-one RNNs, we set $c=10$. {For panel \textbf{A}, rank-one RNNs were identical to those in Fig. \ref{figure3} and had the same initialization procedure. The full-rank RNNs were initialized with $W_{ij} \sim \mathcal{N}(0,N^{-1})$, $b_i = 0$, $(w_{\rm out})_{i}\sim \mathcal{N}(0,N^{-1})$, and $b_{\rm out}\sim \mathcal{N}(0,N^{-1})$. For panel \textbf{B}, the readouts are taken from the full population $\vec x(t)$ across all ranks with $\psi(\cdot)=\sigma(\cdot)$ output nonlinearity. The networks were initialized similar to full-rank RNNs, except for $M_{ij}\sim \mathcal{N}(0,N^{-1})$ and $N_{ij}\sim \mathcal{N}(0,N^{-1})$.} }
    \label{figure6}
\end{figure*}

To start with, we equip the full-rank RNNs with readouts
\begin{equation}
    \hat o(t) = \psi(\vec w_{\rm out}^T \vec x(t) + b_{\rm out}),
\end{equation}
where $\psi(\cdot)$ is a fixed nonlinearity: sigmoid (DA and DMTS tasks) or tanh (DCD tasks); {and readout is now taken from the full population via $N$-dimensional weights $\vec w^{\rm out}\in \mathbb R^N$ with $b^{\rm out} \in \mathbb R$ being the output bias}. {For an autonomous RNN, whether fixed, slow, and ghost points exist depends on the subset of parameters $\vec \theta = \{\vec W,\vec b\}$. To identify such speed minima,} we analyzed autonomous dynamics by setting all inputs to zero, as task inputs are sparse. Following \cite{sussillo2013opening}, we define an RNN-specific energy function as,
\begin{equation}\label{eq:en_min}
     s_{\vec \theta}(\vec x) = \frac{1}{2} \sum_{i=1}^N \left[-x_i + \tanh\!\left( \sum_{\alpha=1}^N W_{i\alpha} x_\alpha + b_i \right) \right]^2,
\end{equation}
which we minimize over $\vec x$ while holding $\vec \theta$ fixed. Since this nonconvex optimization can yield multiple minima, we initialize $\vec x$ with neural activities observed during normal network operation and perform minimization using a PyTorch-based model. A detailed implementation is provided in the shared codebase ({see the data and code availability statement at the end}). Using this procedure, we reverse-engineered the inner workings of two representative RNNs as they performed DCD and DMTS tasks, paralleling our discussion in Fig. \ref{figure4} for low-rank RNNs. 

For the full-rank RNN trained on the DCD task (Fig. \ref{figure5}\textbf{A-B}), training dynamics showed sharp loss decreases within the first $\sim 200$ epochs, accompanied by oscillatory gradients. Extracting speed minima from neural trajectories revealed the emergence of two new fixed points as the underlying state-space phenomenon during this transition. This bistable dynamical system, similar to the low-rank RNN in Fig. \ref{figure4}\textbf{A-D}, enabled the network to map input cues to the correct outputs. At later stages, refinements in learning introduced a ghost point, identified through the eigenvalue spectrum of the Jacobian computed from Eq. \eqref{eq:en_min}. {Specifically, this Jacobian exhibited one (approximately) zero eigenvalue and otherwise negative values, in line with our definition of a ghost in Section \ref{sec:theory}\textbf{A}.} This ghost prolonged trajectories through a transient slow region for one cue, while for the other trial type, the RNN delayed its response by curving trajectories in the high-dimensional state space (blue vs. orange lines in the final two panels of Fig. \ref{figure5}\textbf{B}). Thus, both the ghost-induced bottleneck and the curved neural trajectory enabled the network to suppress premature responses and generate appropriately timed outputs, with the former closely paralleling the progression of learned mechanisms observed in low-rank RNNs (Fig. \ref{figure4}\textbf{A-D}).

For the full-rank RNN trained on the DMTS task (Fig. \ref{figure5}\textbf{C-D}), training dynamics again showed abrupt loss transitions and oscillatory gradients. By epoch 1001, the network had developed a ghost in its state space that indiscriminately delayed responses for all trials (Fig. \ref{figure5}\textbf{D}; \textit{first two panels}). Trajectories around this ghost closely resembled the curved shapes observed in low-rank RNNs (Fig. \ref{figure4}\textbf{E-H}). As training progressed, a slow point also emerged: a local minimum of the energy function that, unlike a ghost, can have positive eigenvalues in its Jacobian spectrum (Fig. \ref{figure5}\textbf{D}, \textit{third panel}). Yet, with continued training, {the slow point was not found and the network introduced additional fixed points.} These fixed points, together with the existing ghost, were fine-tuned to gate trajectories during the delay period before the arrival of the second cue, eventually yielding a correctly timed match vs. no-match response (Fig. \ref{figure5}\textbf{D}, \textit{final panel}). Thus, the full-rank RNN trained on the DMTS task relied on a combination of fixed points, slow points, and ghosts to {learn} working-memory computations, paralleling the mechanisms observed in low-rank RNNs (Fig. \ref{figure4}\textbf{E-H}).

Overall, while earlier work established that full-rank RNNs rely on speed minima to perform working-memory tasks \cite{sussillo2013opening}, our results show that ghosts, as defined in Section \ref{sec:theory}A, also emerge in full-rank networks and play roles analogous to those in low-rank counterparts. There is, however, a key difference in interpretability: low-rank RNNs permit far more direct, analytical insights into the underlying mechanisms ({Figs. \ref{figure3} and \ref{figure4}}), whereas in full-rank networks the same structures are present but harder to disentangle and require numerical analysis (Fig. \ref{figure5}).

\subsection{Partial stabilization of critical learning rates with loosely constrained ranks}

In our canonical model of ghost mechanism, which was trained to perform the DA task (Section \ref{sec:theory}), we identified a critical learning rate above which learning failed; lower rates led to eventual convergence, but rates just below this threshold yielded the fastest early progress (Fig.~\ref{figure2}\textbf{D}). Consistent with this, {when we studied rank-one RNNs with outputs defined as in Eq. \eqref{eq:lat-red} that utilized ghost mechanism to solve this task (see Fig.~\ref{figure3}), we found} the same qualitative pattern (Fig. \ref{figure6}\textbf{A}, \textit{left}). Full-rank RNNs, in contrast, have access to more expressive resources (\textit{e.g.}, additional \textit{loosely constrained} ranks), which could in principle stabilize training dynamics. As before, full-rank RNNs continued to exhibit a critical learning rate (Fig.~\ref{figure6}\textbf{A}, \textit{right}), but the added ranks did substantially stabilize {and improve} the learning dynamics: the critical learning rate increased by nearly two orders of magnitude, and rates near this value became optimal for sustained training (reminiscent of the \textit{catapult mechanism} {studied in a feedforward context \cite{lewkowycz2020large}).}

We next examined learning curves across different rank configurations (Fig. \ref{figure6}\textbf{B}) and found that increasing the number of trainable ranks in $\vec W$ consistently resulted in learning with fewer epoch. Notably, full-rank RNNs exhibited the loss jump up to two orders of magnitudes faster {compared to their rank-one counterparts} {(Fig. \ref{figure6}\textbf{B}; initial loss jumps happen rapidly in all cases, note the log-scale of the x-axis)}. {While some of the observed improvements in training speed may arise from changes in the effective initialization induced by varying rank constraints (which is an interesting direction for future work on initialization strategies for low-rank RNNs), we found that higher-rank RNNs were substantially less likely to become trapped in no-learning zones, particularly at larger learning rates (Fig. \ref{figure6}\textbf{B}).} Despite this stabilization, full-rank RNNs still {entered and often remained in} no-learning zones and occasionally encountered oscillatory minima (Fig. \ref{figure6}\textbf{B}). Thus, while loosely constrained ranks helped mitigate learning instabilities, they did not fully eliminate them. In summary, full-rank RNNs performing the DA tasks can still have oscillatory minima and exhibit a critical learning rate beyond which learning ceases, though adding ranks increases the range of effective learning rates and improves the training stability.

\begin{figure}[!h]
    \centering
    \includegraphics[width=\linewidth]{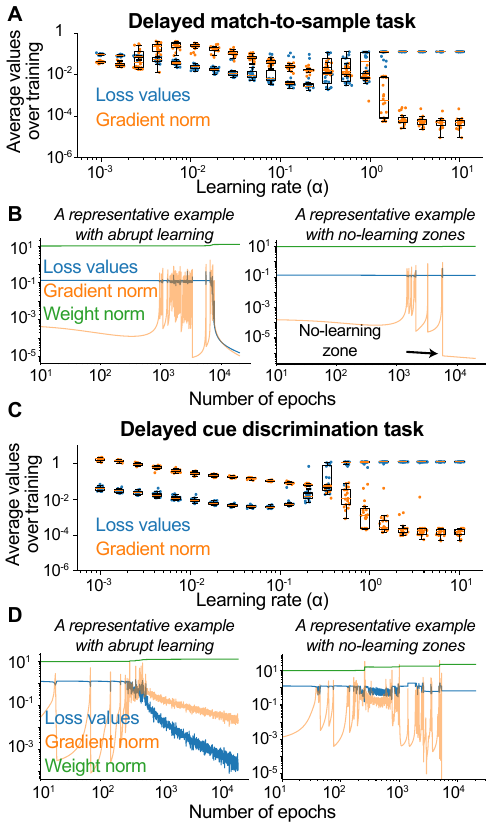}
    \caption{\textbf{Full-rank RNNs exhibit both oscillatory gradients and no-learning zones.} \textbf{A-B} Full-rank RNNs trained on DMTS tasks. \textbf{A} Loss function values and Frobenius norms of recurrent weight gradients as a function of learning rate. {Each dot represents a training run with a different seed ($N_{\rm epochs} = 20,000$), averaged over all epochs. \textbf{B} {Learning} dynamics for two example runs, $\alpha \approx 1.44$. {The green solid line corresponds to the Frobenius norm of the recurrent weights during training.} \textbf{C-D} Same as in \textbf{A-B}, but for full-rank RNNs performing DCD tasks. For panel \textbf{D}, we used two representative runs trained with the learning rate $\alpha \approx 0.34$.} Parameters: $\tau = 10ms$, $\Delta t = 5ms$, $T_{\rm inp}= 30ms$, $T_{\rm delay}=80ms$, $T_{\rm resp}=50ms$, and $N=100$ neurons. {All firing rates were initially zero and no noise was added during time evolution. RNN parameters were initialized following the same procedure as in Fig. \ref{figure5} and training ensued with gradient descent (each batch contains one trial per type).}}
    \label{figure7}
\end{figure}

{

\subsection{Oscillatory gradients and no-learning zones in full-rank RNNs performing working memory tasks}

Encouraged by the results on the DA tasks, we next considered full-rank RNNs trained to perform DMTS and DCD tasks. Notably, unlike the DA task (Fig. \ref{figure3}), RNNs often blend in ghosts within other strategies to solve these tasks (Figs. \ref{figure4} and \ref{figure5}) and therefore it is, so far, an unanswered question whether these other mechanisms make RNNs immune to the no-learning zones. Here, we test this explicitly by training RNNs to perform these tasks across a range of learning rates.

We started with the DMTS task, and
}
as before, observed the emergence of a critical learning rate in this task (Fig. \ref{figure7}\textbf{A}), beyond which learning failed. Interestingly, networks trained on this task showed abrupt improvements in their loss functions, often accompanied by large (and sometimes oscillatory) changes in the gradient norm (Fig. \ref{figure7}\textbf{B}). Moreover, in line with Fig. \ref{figure6}, these full-rank RNNs displayed minimal oscillations near the minima, and in some cases, these oscillations were absent altogether (Fig. \ref{figure7}\textbf{B}, \emph{left}). At high learning rates, gradients remained small even when averaged over the full training window, indicative of no-learning zones (Fig. \ref{figure7}\textbf{A}). When examining individual training instances, we indeed found no-learning zones, in which gradients were effectively zero (Fig.~\ref{figure7}\textbf{B}). {Next, we performed the same experiment with full-rank RNNs performing DCD tasks, which qualitatively agreed with our findings from the DMTS task (Fig. \ref{figure7}\textbf{C-D}).}  

In standard deep learning, vanishing gradients typically originate from saturating neuron-wise activation functions, and are addressed using piecewise-linear activations such as ReLU \cite{glorot2011deep}. However, analysis of the RNN exhibiting a no-learning zone in Fig. \ref{figure7}\textbf{B} (data available via Zenodo, {refer to code designated to Figure \ref{figure7}}) revealed that most firing rates lay outside the saturation regime of $\tanh(\cdot)$, that is, the hidden unit activations were not saturated. Thus, the vanishing gradients observed here stem from a different source: As dynamical systems, RNNs are sensitive to small parameter changes that can qualitatively alter their internal dynamics. Confident outputs inherited from earlier epochs can trap the network in a state where the output remains confidently incorrect, blocking further learning. (See, for example, Fig. \ref{figure3}\textbf{B}, where post-bifurcation dynamics remain stuck in negative $\kappa$ values, yielding a confident zero output.) 

{ Overall, our findings here reveal that in full-rank RNNs performing complex behavioral tasks, critical learning rates and no-learning zones emerge as generic consequences of how dynamical systems navigate optimization landscapes. Notably, no-learning zones arise from dynamical failure modes rather than the architectural or numerical limitations typically encountered in deep learning.}

\subsection{Lowering spurious confidence encourages recovery from no-learning zones}

\begin{figure*}
    \centering
    \includegraphics[width=\textwidth]{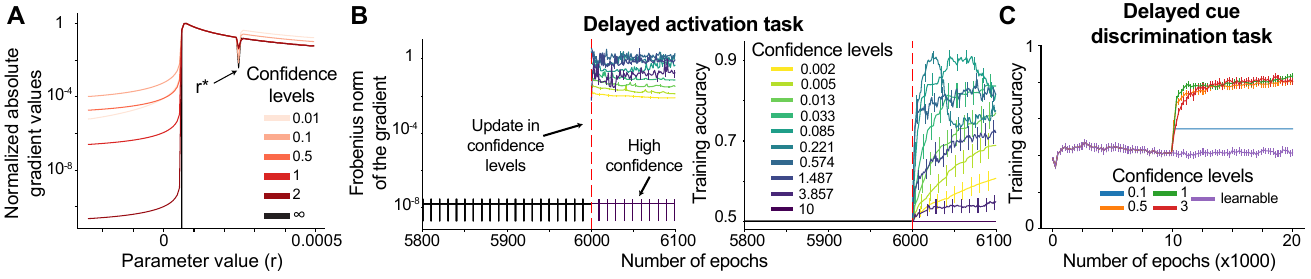}
    \caption{\textbf{Lowering the spurious confidence allows recovering from the no-learning zone.} \textbf{A} To illustrate how lowering output confidence enables gradients to build beyond the point of no return in the ghost mechanism, we plotted the gradients of the canonical model with respect to its scale parameter $r$, otherwise using the same parameters as in Fig.~\ref{figure2}\textbf{A}. Notably, excessively low confidence leads to incorrect outputs, indicating the existence of an optimal confidence level for training (here, around $c \approx 0.1$). For infinite confidence, the gradient is exactly zero for $r<r^*/4$.  \textbf{B} To test the confidence lowering in practice, we trained 100 rank-one RNNs with a confidence $c = 10$ and the learning rate $\alpha = 0.02$ for 6000 epochs, otherwise using the same parameters as in Fig. \ref{figure3}. Out of 100, 74 RNNs learned the task at some point, whereas 48 of them entered the no-learning zone (quantified as having training accuracy of $\leq 0.5$ for the last $50$ epochs). We further trained these RNNs after lowering the confidence levels, which allowed them to recover. Left panel plots the Frobenius norm of the flattened gradient over all parameters, whereas right panel plots the training accuracy as a function of number of epochs. Solid lines: means. Error bars: s.e.m. over 48 networks that were in the no-learning zones within the epochs $[5950,6000]$. {Training accuracy is defined as the agreement between discretized network outputs and target outputs (cf. Fig. \ref{figure5}).} \textbf{C} To generalize the confidence lowering to higher-rank RNNs, we trained full-rank RNNs on DCD tasks with {$\alpha = 0.1$}, and compared the training dynamics of fixed vs. learnable output confidence. During the first $10{,}000$ epochs, $c$ is trainable in both cases. Solid lines: means. shaded areas: s.e.m. over 100 networks. Parameters {for \textbf{C}}: $\tau = 10ms$, $\Delta t = 5ms$, $T_{\rm inp}= 10ms$, $T_{\rm delay}=50ms$, $T_{\rm resp}=50ms$, and $N=100$ neurons.  {Firing rates were initialized with $x_i\sim \tanh( \mathcal{N}(0,0.1^2))$, random  noise (with s.d. $10^{-3}$)} added to the RNN at every time step, and training with gradient descent with 20 trials (10 per cue) using Adam optimizer \cite{kingma2014adam}. {RNNs were initialized following the same procedure as in Fig. \ref{figure5}.} }
    \label{figure8}
\end{figure*}

Given the inevitability of the no-learning zones, which appear whenever learning rates exceed critical values, we next set out to identify {a} mechanism through which they {can} arise. In earlier literature, vanishing gradients were extensively studied and attributed to the repeated multiplication of Jacobian matrices in saturated regimes of nonlinearities, which drives gradients toward zero \cite{pascanu2013difficulty}. A natural question is whether the no-learning zones observed here reflect the same phenomenon, or instead emerge from a distinct mechanism. 

To answer this question, we return to our canonical model solving the DA task (Fig.~\ref{figure2}) and analyze why no-learning zones arise. Recall that with infinite confidence ($c=\infty$), the model outputs $0$ or $1$ depending on whether the state variable $\kappa(t)$ exceeds a threshold $\bar \kappa \gg 1$. For a given $T$, the optimal solution scales as $r^{*} \propto T^{-2}$ (Eq.~\eqref{eq:error}), yielding outputs of $0$ for the first $T$ time steps and $1$ thereafter. As discussed in Section~\ref{sec:theory}, this scaling not only complicates training for large $T$, but also places $r^{*}$ close to the value $\tilde{r} := r^{*}/4$, where the output remains identically zero throughout the full $2T$ duration. For $r < \tilde{r}$, variations in $r$ no longer affect the output, which stays zero despite infinitesimal parameter changes. The loss function therefore becomes flat with zero local gradient.  

If oscillations around the optimum or stochastic fluctuations drive the system into this regime, learning halts, independent of whether neural activities are saturated. The resulting no-learning zones are thus distinct from the classical vanishing gradient problem: they arise not from architectural properties such as Jacobian contraction in saturated nonlinearities, but from the structure of the underlying dynamical system, where small parameter shifts (\textit{e.g.}, from $\tilde r+\delta$ to $\tilde r-\delta$ for some small $\delta$; black line in Fig.~\ref{figure8}\textbf{A}) can induce qualitatively different outputs. This abrupt change reflects the infinite confidence of the canonical model. By contrast, fixing confidence to finite values during training preserves small but nonzero gradients even within the no-learning zone, allowing eventual recovery (Fig.~\ref{figure8}\textbf{A}). In other words, the no-learning zones originate from the overconfidence of dynamical models in their outputs: small parameter changes in the dynamics can globally alter predictions, leaving training trapped in incorrect yet overly confident solutions.  

We next tested this prediction in practice with rank-one RNNs trained on the DA task, for which we have already established no-learning zones for large enough learning rates (Fig. \ref{figure3}). Specifically, we asked whether artificially lowering confidence levels can help them escape no-learning zones. To test this, we returned to rank-one RNNs performing DA tasks and lowered their output confidence ($c$, Eq. \eqref{eq:lat-red}) after they were stuck in no-learning zones (Fig.~\ref{figure8}\textbf{B}). As expected, we found that reducing confidence allowed the RNNs to become unstuck and re-learn the task (Fig.~\ref{figure8}\textbf{B}). This aligns with the predictions of the canonical model, in which lower confidence values permit small but nonzero gradients to form beyond the point of no return (Fig.~\ref{figure8}\textbf{A}).

The results from the canonical model and rank-one RNNs suggest a general principle, which we next tested in full-rank RNNs performing DCD tasks. To set and fix the output confidence level in full-rank RNNs, we defined the readout as follows:
\begin{equation}
    \hat o(t) = \tanh\left( c  \frac{\vec w_{\rm out}^T \vec x(t) + b_{\rm out} }{||\vec w_{\rm out}||_1} \right),
\end{equation}
where the division by $||\vec w_{\rm out}||_1$ ($\mathcal{L}_1$ norm) introduces scale invariance to $\vec w_{\rm out}$. This has effectively fixed the confidence levels of the $\tanh(\cdot)$ nonlinearity, similar to Eq.~\eqref{eq:lat-red} for the rank-one RNNs. 

To better approximate realistic training conditions, we trained full-rank RNNs using batches and the ADAM optimizer \cite{kingma2014adam}, introduced noise into the firing rates, and randomized the initial state across trials. With this setup, we trained full-rank RNNs using a high learning rate that could produce no-learning zones (Fig. \ref{figure8}\textbf{C}, purple line). As expected, RNNs trained under this condition failed to improve output accuracy (Fig. \ref{figure8}\textbf{C}). Then, we manually intervened at epoch 10,000 by lowering and fixing the output layer’s confidence (Fig. \ref{figure8}\textbf{C}). Following this intervention, output accuracy increased compared to the baseline condition (Fig. \ref{figure8}\textbf{C}), { in some cases even reaching perfect performance. However, excessively low confidence values again prevented learning from being sustained in the long run.}

\section{Discussion} \label{sec:discussion}
{
In this work, we introduced a canonical model that provides an analytical description of abrupt learning through the formation of ghost points. This model is not merely illustrative but rather an exact reduction of high-dimensional dynamics near ghosts, recapitulating their learning dynamics in an architecture-agnostic manner. With this model, we derive an analytical scaling law for the critical learning rate beyond which formation of ghosts becomes impossible. While there are likely other mechanisms for learning long-term dependencies \cite{pascanu2013difficulty}, this theoretical discovery provides one important piece of the puzzle for why longer working memory tasks seem to require increasingly fine-tuned optimization, \textit{i.e.}, the solution, and not necessarily the architecture, requires it. We validated these theoretical predictions across both low-rank and full-rank RNN architectures, which involved both the predicted oscillatory gradients and no-learning zones. Furthermore, we identified two practical strategies for stabilizing ghost-based learning: increasing the number of trainable ranks, which partially stabilizes learning at higher learning rates, and reducing output confidence, which mitigates the associated vanishing gradient problem in full-rank networks.

\subsection{Why does ghost mechanism lead to abrupt learning?}

An important aspect of the ghost mechanism is its locality in both dynamics and the scale parameter. Losing sight of this fact, and studying Fig. \ref{figure2}\textbf{A}, may lead to an incorrect interpretation, in which ghosts emerge only within a small range of the scale parameter $r$. Moreover, it is not necessarily clear why $r$ should start from large positive values rather than, say, negative values. The key insight is that we are interested in networks that are randomly initialized, as in Fig. \ref{figure3}. As shown for rank-one RNNs, random initialization is unlikely to have any meaningful fixed point structure in place  \footnote{That is, unless initialization occurs in parameter subspace that leads to chaos in RNNs \cite{sompolinsky1988chaos}, which opens an entirely different research direction}. Instead, as learning progresses, the ghost emerges spontaneously at a specific location (Fig. \ref{figure3}\textbf{A}), and this emergence is rapidly governed as $r$ approaches $O(r^*)$ from a regime with no fixed points at all, \textit{i.e.}, from large $r$ values. In this regime, the gradient rapidly changes from flat to very high values as $r$ approaches $O(r^*)$ following Eq. \eqref{eq:gradient}, causing learning to occur abruptly.

That being said, the loss function in Eq. \eqref{eq:error} describes only the local behavior of the scale parameter when the speed minimum is already formed. Based on the success of training across various conditions and initial seeds (Fig. \ref{figure3}), there are likely many parameter configurations in high-dimensional space that can map current RNN parameters to a local ghost mechanism, \textit{i.e.}, following Eq. \eqref{eq:eff_r}. On the other hand, the large gradients emerging after entering a parameter region governed by the ghost mechanism are simultaneously both the blessing and curse of optimization. For learning rates that scale as $T^{-4}$ for delay duration $T$ (and therefore vanish when long-term dependencies must be learned), these large gradients can trigger saddle-node bifurcations and eventual entrapment in no-learning zones. This dual nature of the loss function clarifies why the ghost mechanism can lead to both abrupt learning of new capabilities and entrapment in regions of parameter space with vanishing gradients. 

\subsection{A dynamics-first approach to studying learning dynamics}

Much of the machine-learning literature approaches learning dynamics by first selecting specific architectures that exhibit phenomena of interest (\textit{e.g.}, abrupt learning), then analyzing their behavior \cite{eisenmann2023bifurcations,haputhanthri2024why,bahri2020statistical}. However, this architecture-first strategy can limit our ability to extract universal principles of computation and learning. Here, we adopt a \textit{dynamics-first} perspective by focusing on problems solved by dynamical systems, specifically working memory tasks. We find that one common solution, the ghost mechanism, enforces universal constraints on both the time evolution of state variables and the learning dynamics of network parameters. These constraints enabled us to predict diverse phenomena across architectures: the emergence of oscillatory gradients, the existence of critical learning rates that scale with delay duration, and the formation of no-learning zones where optimization fails entirely. By deriving these predictions from the dynamical solution rather than architectural specifics, we obtain results that transfer from our simple canonical model to low-rank and full-rank RNNs.

This success of the dynamics-first approach is particularly relevant for systems neuroscience, where the goal is {often} not to replicate the brain's exact architecture but to understand computational principles that govern neural processing \cite{richards2019deep,yang2019task,mante2013context,dinc2025latent,dubreuil2022role,perich2021inferring,perich2025neural,perich2020rethinking}. When RNNs are trained to match neural recordings or perform behavioral tasks, they serve as \textit{digital twins} that capture functional dynamics, which can be tested causally with experimental manipulations \cite{liu2024encoding}, rather than replicating biological details at the level of individual neurons \cite{das2020systematic}. These models are valuable precisely because they abstract away from implementation details to reveal computational motifs. In this sense, neuroscientists have implicitly embraced a dynamics-first approach, without necessarily articulating the associated mathematics. Our canonical reduction makes this philosophy explicit and precise: by isolating the ghost mechanism, we identify {one} fundamental dynamical structure that {endows finite-dimensional continuous-time dynamical systems with the ability to} maintain information over time, regardless of whether those systems are artificial or biological. 

Indeed, ghost formation itself is directly relevant for computational neuroscience: dynamical bottlenecks, like ghosts shown in Figs. \ref{figure4} and \ref{figure5}, are routinely observed in trained RNNs, \textit{i.e.}, in silico models of neural computation \cite{mante2013context,maheswaranathan2019reverse,dinc2025latent}, and may underlie short-term memory in neural circuits {\cite{koch2024ghost}}. Previous work has typically identified these structures post hoc, using numerical methods to analyze fully trained networks \cite{sussillo2013opening}. Our approach complements this by revealing how such bottlenecks form during training, exposing both their computational utility and the learning instabilities they create. The canonical model in Eq.~\eqref{eq:ghost-prototype} thus provides a conceptual bridge between abstract dynamical systems theory and empirical observations in neuroscience.
{ Specifically, for dynamical systems to learn ghosts, learning rates should decrease as the timescale of learned computations increases. Working memory tasks with variable delay periods provide a natural setting to test this prediction. We anticipate two key behavioral signatures: (i) animals should exhibit lower effective learning rates when trained on longer delay periods, and (ii) interventions that artificially increase learning rates (\textit{e.g.}, through pharmacological manipulation) may paradoxically impair learning by driving network parameters into no-learning zones, resulting in accuracy plateaus long after the intervention ends. Testing these predictions on behaving animals requires observing and modifying correlates of learning rates in brains, towards which promising research efforts have long been devoted in the field \cite{iigaya2018effect,grossman2022serotonin,coddington2023mesolimbic}.}

Finally, this dynamics-first framework also suggests that when neuroscientists observe slow dynamics or persistent activity in neural recordings \cite{rajan2016recurrent,fuster1971neuron,vyas2020computation,honey2012slow}, these phenomena may not simply reflect arbitrary network properties but could arise from fundamental constraints on how general dynamical systems learn to perform temporal computations. The ghost mechanism predicts specific signatures in both dynamics (transient slowdowns near critical points) and learning (oscillatory gradients, critical learning rates), which could be tested in biological learning experiments or in the optimization of RNN models of neural data \cite{dinc2023cornn,valente2022extracting}. Such tests would help determine whether ghost points represent a universal solution that evolution and learning converge upon when faced with the challenge of maintaining information over time, and if so, what other solutions are out there.

\subsection{Ghost signatures persist despite increasing architectural complexity}

Although the ghost mechanism we describe here is fundamentally one-dimensional \cite{strogatz2018nonlinear}, RNNs are known to utilize higher-dimensional slow point manifolds (\textit{i.e.}, geometric analogs of ghost points) to solve memory tasks \cite{sussillo2013opening, ribeiro2020beyond, schmidt2021identifying}. This is an area of growing interest in computational neuroscience \cite{sagodi2024back}, in which high-dimensional state spaces of full-rank RNNs enable a broader repertoire of mechanisms (beyond ghost-like attractors) to construct and sustain such manifolds. This raises the possibility that the ghost dynamics observed in our model may reflect artifacts of low-rank constraints or task simplicity, and not necessarily one central mechanism of abrupt learning associated with delaying network outputs. This is certainly true to some extent, as we have observed that higher dimensions contain new strategies for delaying network outputs (see Figs. \ref{figure4} and \ref{figure5} with the curved neural activities; also refer to \cite{kurtkaya2025dynamical}). Yet, as we show in Fig. \ref{figure5}, full-rank RNNs still do learn ghosts and as we show in Fig. \ref{figure7}, the core insights from our simplified model generalize to the learning dynamics of full-rank RNNs performing complex working memory tasks.

Our progression from simple to complex systems reveals a fundamental trade-off between interpretability and generality. In the canonical model, we achieved complete mechanistic understanding: exact scaling laws, precise conditions for instability, and analytical expressions for critical parameters. Moving to low-rank RNNs (Fig. \ref{figure4}), we could still trace how specific dynamical objects (ghosts, fixed points, slow points) contribute to computation, visualizing the complete phase portrait and identifying each component's functional role. However, in full-rank networks (Fig. \ref{figure5}), our analysis necessarily becomes more limited. While we can identify ghost formations and observe characteristic neural trajectories in principal component projections, the full mechanistic picture remains opaque. This progression mirrors a broader challenge in the field: we often need to trade analytical tractability for biological realism.

This trade-off has shaped how the field of computational neuroscience studies high-dimensional RNNs. Earlier work has already made attempts at characterizing these systems, though even then the search has been for low-dimensional features (as in Fig. \ref{figure5}) that can be studied with known numerical methods \cite{sussillo2013opening}. As evidenced by existing literature, this is not always overly restrictive, since in many cases full-rank RNNs do learn low-dimensional, effectively low-rank, solutions {\cite{valente2022extracting,krause2022operative}}. That being said, characterizing the full range of solutions learned by full-rank RNNs, or predicting which degenerate solution will be learned to solve a particular task in general networks \cite{richards2019deep,huang2024measuring}, remains a major mystery studied across the board. Our work contributes to this effort, {\textit{i.e.}, at least one mechanism, the ghost, has appeared} consistently across this complexity gradient as long as the demands of the tasks lead to ghost formation.

\subsection{Outlook: Broader implications}

Our analysis of ghost mechanisms reveals fundamental constraints on how dynamical systems learn temporal computations, with implications spanning neuroscience, machine learning, and biological learning. For machine learning, our findings illuminate why training RNNs on temporal tasks remains challenging despite architectural advances. Previous work has shown that RNNs often converge to effectively low-rank solutions \cite{schuessler2020interplay,krause2022operative,valente2022extracting}, achieved through two distinct strategies: (1) restricting updates to low-rank components from the outset, as in FORCE learning \cite{sussillo2009generating} and traditional low-rank training \cite{dubreuil2022role,beiran2021shaping,valente2022extracting}, or (2) {allowing unconstrained updates that gradually refine into a low-rank structure during learning (see \cite{balzano2025overview} for recent related ideas in the deep learning literature)}. Our analysis suggests strategy (2) may inherently stabilize learning by providing additional degrees of freedom during training, even when the final solution is low-dimensional. This insight could inform the design of training algorithms that delay rank constraints until later in training while explicitly leveraging high-dimensional parameter spaces early in the training to navigate around no-learning zones before converging to simpler representations.

Looking forward, the ghost mechanism exemplifies how careful analysis of minimal models can yield insights that scale to complex systems. As the field develops increasingly sophisticated architectures for temporal processing, from Transformers \cite{vaswani2017attention} to state-space models \cite{gu2024mamba}, understanding fundamental dynamical constraints becomes ever more critical. {Beyond recurrent systems, feedforward architectures like deep neural networks and large language models exhibit analogous abrupt transitions marking the acquisition of new computational capabilities  \cite{yu2023skill, lubana2024percolation, okawa2023compositional, wei2022emergent, hoffmann2022empirical, power2022grokking}. Whether ghost mechanisms or fundamentally different dynamical structures produce these transitions in non-recurrent systems remains an important open problem. Extending mechanistic analyses of this kind to spiking neural networks represents a further important direction for bridging theoretical understanding with biologically plausible implementations \cite{klos2025smooth,bellec2020solution,pfeiffer2018deep}.} 

The ghost framework demonstrates that some learning challenges arise not from implementation details but from how any dynamical system must reorganize to maintain information over time. Future work should explore whether similar canonical reductions can illuminate other persistent challenges in learning dynamics, ultimately bridging the gap between theoretical understanding and practical advances in both artificial and biological intelligence. In particular, investigating whether critical learning rates, no-learning zones, and oscillatory gradient dynamics emerge in these modern architectures could reveal whether the constraints we identified are truly universal or specific to RNN-like dynamics.

{One particularly compelling case concerns architectures that introduce delayed inputs directly (most notably attention-based ones \cite{vaswani2017attention,niu2021review}). This stands in contrast to recurrent systems, which tend to maintain temporal context  in their internal dynamics (\textit{e.g.}, via ghosts). This leads to a concrete, falsifiable prediction: Modern models may not be subject to the same inverse power-law scaling of critical learning rates with delay duration identified here. More broadly, future work should systematically study the effects of diverse heuristics in the machine learning literature (\textit{e.g.}, alternative loss functions, regularization strategies, and choices of nonlinearity and model architecture), which may reshape the optimization landscape in ways that interact nontrivially with ghost formation and the associated instabilities.}

Finally, our findings have implications for future research in physics, biophysics, and neuroscience. First, dimensions in parameter space that have little effect on the system’s output or loss are universally observed in many physical models \cite{gutenkunst2007universally}. Our findings suggest that one reason for their existence may be the need to stabilize the learning dynamics of few well-constrained dimensions. Second, in the traditional deep learning paradigm, the confidence levels could be optimized, not artificially controlled, and therefore the training process would be susceptible to no-learning zones. We presented a potential remedy in Fig. \ref{figure8}, \textit{i.e.}, different mechanisms may be crucial for learning (practicing) versus inference (performing). This observation underscores the importance of incorporating mechanisms beyond loss minimization (\textit{e.g.}, simulated annealing \cite{kirkpatrick1983optimization}) to modulate system behavior when learning stalls in neural networks, and mirrors seminal work on adult zebra finch, where males exhibit variability during practice, but perform with precision and reduced neural variability in the presence of a female \cite{hessler1999social,kao2008neurons}. The parallel suggests that biological systems may have evolved similar strategies to avoid dynamical traps during learning, providing a concrete prediction about how neural circuits might modulate confidence during skill acquisition.
}

\subsection*{Data and code availability}

The code and data used to support the findings of the study are both available at \href{https://doi.org/10.5281/zenodo.13686989}{https://doi.org/10.5281/zenodo.13686989}. For additional details on the training and experimental parameters, readers can also refer to the Github codebase: \href{https://github.com/fatihdinc/ghost-mechanism}
{https://github.com/fatihdinc/ghost-mechanism}.

\subsection*{Acknowledgements}
We would like to thank Dr. Marta Blanco-Pozo, Dr. Saeed Ahmed Khan, Dr. Sarah Kushner, and  Dr. Daniel Kunin for their insightful comments on the manuscript, Dr. Nina Miolane, Abby Bertics, and members of the Geometric Intelligence Lab at UC Santa Barbara for their helpful feedback on an earlier version of the project, and Dr. Boris Shraiman for fruitful discussions on the loosely constrained parameters and learning in high-dimensional dynamical systems. MJS gratefully acknowledges funding from the Simons Collaboration on the Global Brain, the Vannevar Bush Faculty Fellowship Program of the U.S. Department of Defense, and Howard Hughes Medical Institute. FD receives funding from Stanford University's Mind, Brain, Computation and Technology program, which is supported by the Stanford Wu Tsai Neuroscience Institute. EC and BK's internships were supported in part by a grant from the Feldman-McClelland Open-a-Door fund of the Pittsburgh Foundation. BK thanks the Impact Scholarship and Research Scholarship for Critical Thinkers programs from the Bridge to Turkiye Funds for supporting his visit to Stanford University. Some of the computing for this project was performed on the Sherlock cluster. We would like to thank Stanford University and the Stanford Research Computing Center for providing computational resources and support that contributed to these research results. This research was supported in part by grant NSF PHY-2309135 and the Gordon and Betty Moore Foundation Grant No. 2919.02 to the Kavli Institute for Theoretical Physics (KITP).

\appendix
\setcounter{figure}{0}
\renewcommand{\thefigure}{S\arabic{figure}}
\renewcommand{\theHfigure}{S\arabic{figure}}

\section{Methodological details for reproducibility} \label{app}

\subsection{Task details}

In this paper, we investigate three key tasks to understand the nature of the ghost mechanism and evaluate its generalizability across a spectrum of RNNs, from one-rank to full-rank, performing behavioral tasks. To ensure clarity and reproducibility, we describe all three tasks in detail in this section.

\emph{Delayed-activation} task is the first of the three tasks examined using both the toy model and RNNs. This task consists of two distinct intervals: the \textit{delay interval} ($T_{\text{delay}}$) and the \textit{response interval} ($T_{\text{resp}}$). Unlike the other tasks studied in this work, the DA task does not involve any input signal. Instead, the desired output is defined by the following piecewise function $o(t)$:

\begin{equation}
   o(t) = 
   \begin{cases}
        0 & \text{if } t \in \{ T_{\text{delay}} \} \\
        1 & \text{if } t \in \{ T_{\text{resp}} \} \\
   \end{cases}  
\end{equation}

\emph{The delayed match-to-sample} (DMTS) task involves both input, $u(t)$, and output, $o(t)$, signals. This task consists of four distinct periods: the \textit{input interval} ($T_{\text{inp}}$), \textit{delay interval} ($T_{\text{delay}}$), \textit{match interval} (also $T_{\text{inp}}$), and \textit{response interval} ($T_{\text{resp}}$). The task features two cue periods: the input and match intervals. After receiving the input cue (randomly chosen from $\{-1, 1\}$) during the input interval, the model must retain this information through the delay interval. During the match interval, a second cue is presented. The model must determine whether this cue matches the original input. In the response interval, the model should output $1$ if the cues match, and $0$ otherwise. Formally, the input and output signals are defined as follows:

\begin{equation}
   u(t) = 
   \begin{cases}
        \pm 1 & \text{if } t \in \{ T_{\text{inp}}, T_{\text{match}} \} \\
        0 & \text{if } t \in \{ T_{\text{delay}}, T_{\text{resp}} \} \\
   \end{cases} 
\end{equation}
\begin{equation}
o(t) = 
   \begin{cases}
        0 & \text{if } t \in \{ T_{\text{inp}}, T_{\text{delay}}, T_{\text{inp}} \} \\
        1 & \text{if } t \in \{ T_{\text{resp}} \} \text{ and } u_{\text{inp}} = u_{\text{match}} \\
        0 & \text{if } t \in \{ T_{\text{resp}} \} \text{ and } u_{\text{inp}} \neq u_{\text{match}} \\
   \end{cases}    
\end{equation}

\emph{The delayed cue discrimination} (DCD) task also involves both an input, $u(t)$, and an output, $o(t)$, signal. However, unlike the DMTS task, it does not include a match interval, only the input interval serves as the cue presentation. The objective of the task is to retain the input cue through the delay period and produce the corresponding response afterward. Formally, the input and output signals are defined as follows:
\begin{equation}
    u(t) = 
   \begin{cases}
        \pm 1 & \text{if } t \in \{ T_{\text{inp}} \} \\
        0 & \text{if } t \in \{ T_{\text{delay}}, T_{\text{resp}} \} \\
   \end{cases}
\end{equation}
\begin{equation}
   o(t) = 
   \begin{cases}
        0 & \text{if } t \in \{ T_{\text{inp}}, T_{\text{delay}} \} \\
        u_{\text{inp}} & \text{if } t \in \{ T_{\text{resp}} \} \\
   \end{cases} 
\end{equation}
Here $u_{\rm inp}$ refers to the input presented during the input window at the particular trial.

\subsection{Analyses with RNNs} 
There are several important distinctions between the trained RNNs and our toy model. First, for large $\kappa$, the time evolution equations asymptotically becomes:
\begin{equation}
    \tau \dot \kappa(t) \to -\kappa(t) + O(1), \quad |\kappa| \to \infty,
\end{equation}
where $O(1)$ refers to the $\kappa$ independent saturation term due to the saturation of $\tanh(.)$ non-linearities. Thus, unlike the toy model, RNN equations cannot generate a global ghost point, as there is always at least one fixed point to accommodate $f(\kappa;\vec m,\vec n,\vec b) \to -\kappa$ limit as $\kappa \to \pm \infty$. Instead, as we show in Fig. \ref{figure3}, the ghost point emerges locally. 

Second, the RNN equations have a characteristic time-scale, $\tau$, with respect to which we perform an Euler discretization and design the delay period $T$. Specifically, we pick $\Delta t = 5ms$ such that $ \Delta = \Delta t/\tau = 0.5$ and the time evolution equation becomes:
\begin{equation} \label{eq:rnn_evolution}
    \vec x[t+\Delta t] = (1-\Delta )\vec x[t] + \Delta  \tanh(\vec z[t]),
\end{equation}
{where $z[t] = \vec W\vec x[t] + \vec W^{\rm in} \vec u[t]+ \vec b + \vec \epsilon$.}

A rank-one RNN, despite being characterized by a single dynamical variable, has $O(N)$ parameters. This means there are far more knobs to turn, effectively creating redundancy and loosely constrained parameters. Thus, we do not expect the critical learning rate in RNNs to follow Eq. \eqref{eq:critical}, as the RNNs are more expressive than the toy model. For example, they can learn to change the curvature of the ghost point as shown in Fig. \ref{figure3}\textbf{A}. But, as shown in Fig. \ref{figure6}\textbf{A}, we did recover the same trend between the learning rates and the final loss function values.

For our analysis with increasing ranks in Fig. \ref{figure6}, we used the RNNs described by Eq. \eqref{eq:rnn_eq}, but with the following output:
\begin{equation}\label{eq:readout}
    \hat o(t) = {\sigma(\vec w_{\rm out}^T \vec x(t) + b_{\rm out})},
\end{equation}
where $w_{\rm out} \in \mathbb R^N$ and $b_{\rm out} \in \mathbb R$ are trainable parameters. {Unlike the rank-one RNN settings in earlier figures}, this choice allows to train the confidence parameter ($c$ for the previous models) and the threshold value ($\kappa^*$ or $x^*$ for previous models). We enforce the rank constraints by defining $\vec M \in \mathbb R^{N\times K}$ and $\vec N\in \mathbb R^{K\times N}$ such that
\begin{equation}
   \vec W = \vec M\vec N
\end{equation}
is {at most a rank-K matrix}. We train these networks by first computing the forward evolution using the same discretization procedure explained in Eq. (\ref{eq:rnn_evolution}) and performing stochastic gradient descent on the discretized version of the loss function in Eq. \eqref{eq:loss_original}.

For our analyses on working memory tasks, presented in Figs. \ref{figure5}, \ref{figure7} and Fig. \ref{figure8}, we used two distinct readouts for full-rank RNNs. For the DMTS task, we used the same network output as in Eq. (\ref{eq:readout}). For the DCD task, since the output is contained within $[-1,1]$, we used a $\tanh(\cdot)$ nonlinearity in the readout layer in Eq. (\ref{eq:readout}). For Fig.  \ref{figure4}, we instead computed the readouts from the two-dimensional latent variables as opposed to the neural activities, { with the task-specific output non-linearities as described above.}

\end{document}